%% file: paper.tex
\RequirePackage{amsmath,amssymb,amsthm}
\documentclass[]{bytedance_seed}

\input{math_commands.tex}

\usepackage{algorithm}
\usepackage{algpseudocode}
\usepackage{enumitem}

\newtheorem{theorem}{Theorem}
\newtheorem{lemma}{Lemma}

\newtheorem{proposition}{Proposition}
\theoremstyle{remark}

\theoremstyle{plain}
\newcommand{\diag}{\operatorname{diag}}

\newcommand{\rank}{\operatorname{rank}}

\allowdisplaybreaks

\newcommand{\Method}{\textsc{Lare}}

\title{Beyond Data Schedule: Fast LLM Pretraining on Data Mixtures by Loss-Reweighting}

\author[1,2,*]{Jiazheng Li}
\author[1]{Huaqing Zhang}
\author[1]{Jing Xu}
\author[2]{Tianle Cai}
\author[1, \dagger]{Jingzhao Zhang}

\affiliation[1]{Tsinghua University}
\affiliation[2]{ByteDance Seed}

\contribution[*]{Work done at ByteDance Seed}
\contribution[\dagger]{Corresponding authors}

\newcommand{\agd}{ActGD}

\abstract{\input{sections/00_abstract/abstract}}

\date{\today}
\correspondence{
Jiazheng Li at \email{foreverlasting1202@outlook.com}\\
Jingzhao Zhang at \email{jingzhaoz@mail.tsinghua.edu.cn}}

\begin{document}
\maketitle

\input{sections/01_introduction/section}
\input{sections/02_phenomenon/section}
\input{sections/03_theory_method/section}
\input{sections/04_experiments/section}
\input{sections/06_conclusion/section}



\subsection*{Ethics statement}

This work studies domain-level loss reweighting for language model pretraining. Models trained with this approach may inherit biases, harmful content, and privacy risks from their training data. Reweighting domain losses may also change the influence of these properties, which is not evaluated in this study. Our reported improvements concern downstream task performance and should not be interpreted as evidence of improved safety or fairness. Models developed using this approach require appropriate safety evaluation before deployment.

\subsection*{Reproducibility statement}

To ensure reproducibility, we provide the code, dataset and model in the supplementary materials and anonymous github \url{https://anonymous.4open.science/r/Fast-LLM-Pretrain-on-Data-Mixture-by-Loss-Reweighting-F7F5}. In the README.md file included with the code, we present a step-by-step guide for reproducing our results.


\clearpage
\bibliographystyle{plainnat}
\bibliography{main}

\clearpage
\beginappendix
\input{sections/05_related_work/section}
\clearpage
\input{sections/90_appendix_theory/section}

\clearpage
\input{sections/91_appendix_data/section}
\clearpage
\input{sections/93_appendix_details/section}
\clearpage
\input{sections/94_appendix_scaling/section}
\clearpage
\input{sections/96_appendix_agd/section}

\end{document}

%% file: math_commands.tex
\usepackage{amsmath,amsfonts,bm}

\def\eqref#1{equation~\ref{#1}}

\def\1{\bm{1}}

\DeclareMathAlphabet{\mathsfit}{\encodingdefault}{\sfdefault}{m}{sl}
\SetMathAlphabet{\mathsfit}{bold}{\encodingdefault}{\sfdefault}{bx}{n}

\newcommand{\E}{\mathbb{E}}

\newcommand{\R}{\mathbb{R}}

\newcommand{\Var}{\mathrm{Var}}

\DeclareMathOperator*{\argmin}{arg\,min}

%% file: sections/00_abstract/abstract.tex
Large language models often acquire specialized capabilities through successive pretraining stages, but these gains can come at the cost of forgetting earlier knowledge. In this work, we investigate whether such trade-offs can be improved by changing how strongly each domain contributes to learning while keeping its data exposure fixed. We find that, in controlled knowledge--mathematics experiments, loss reweighting produces a single checkpoint that surpasses the best knowledge and reasoning scores attained along either stage trajectory. This reveals an independent control that data-mixture ratios alone do not capture. Building on this finding, we propose \Method{} (Loss-Adaptive Reweighting under fixed Exposure), which automatically adjusts domain weights from running loss statistics. Motivated by a theoretical analysis of gradient noise, its normalized inverse-square-root rule balances relative loss scales without changing the sampled mixture or requiring additional runs to search for weights. Across knowledge, mathematics, and code, \Method{} improves all five evaluated benchmarks for a 1.5B model trained on 240B tokens, averaging {\color{green!50!black}+7.84 pp} over a matched direct mixture, including gains of {\color{green!50!black}+17.06 pp} on GSM8K and {\color{green!50!black}+9.00 pp} on MATH. These findings demonstrate that adjusting domain loss scales can improve joint capabilities without additional domain exposure.

%% file: sections/01_introduction/section.tex
\begin{figure*}[ht]
  \centering
  \includegraphics[width=0.7\textwidth]{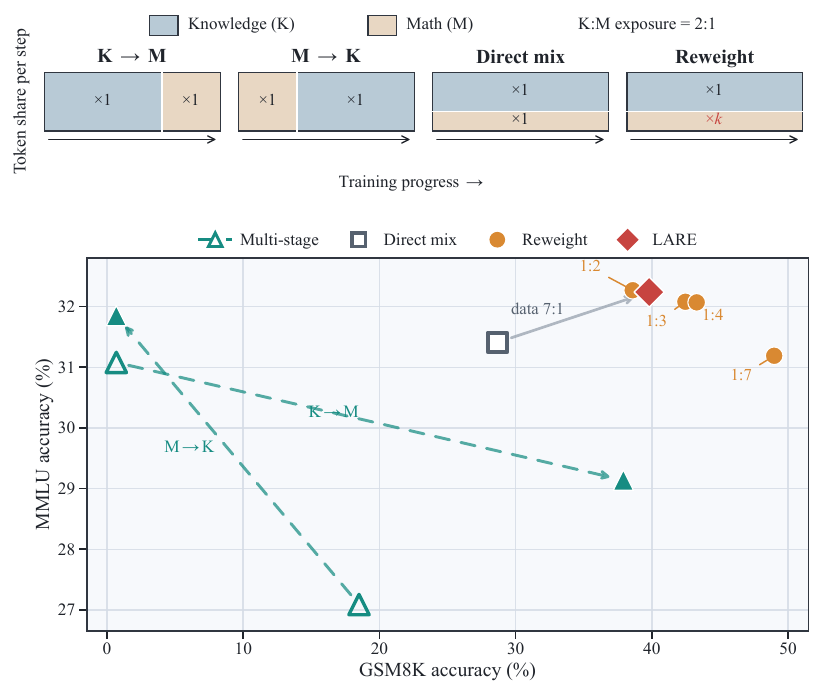}
  \caption{\small\textbf{Loss reweighting, rather than additional domain exposure, yields a joint improvement in knowledge and mathematical reasoning.} \textbf{Top:} colored areas show cumulative Knowledge:Math exposure; the \(2{:}1\) schematic keeps both domains visible. Labels \(\times1\) and \(\times k\) denote relative loss multipliers: direct mixing weights both domains equally, whereas reweighting increases the mathematics loss. \textbf{Bottom:} under matched \(7{:}1\) exposure, open triangles mark first-domain peaks and filled triangles mark selected post-switch checkpoints, while the square, circles, and diamond denote direct mixing, fixed coefficients, and \Method{}. Fixed \(1{:}2\) weighting and \Method{} exceed the coordinate-wise stage maxima; \Method{} reaches this region online without changing exposure.}
  \label{fig:teaser-pareto}
  \vspace{-0.4cm}
\end{figure*}

\section{Introduction}
\label{sec:introduction}

Large language models are increasingly expected to master multiple capabilities simultaneously---from broad world knowledge to mathematical reasoning, code generation, and instruction following. Meeting these demands has pushed pretraining beyond a single stationary data distribution: practical recipes now begin with broad-coverage corpora and subsequently introduce midtraining or continued-pretraining stages enriched with curated, domain-specialized data~\citep{gururangan2020dont,walsh2025olmo2,liu2025midtraining,lu2024mathcoder2}. Yet whether successive stages are the best way to combine multiple domains remains an open question. Emphasizing a new domain can rapidly improve the corresponding capability, but these gains often come at the cost of forgetting what was learned earlier~\citep{ke2023continualpretraininglanguagemodels,yildiz2025continual,marek2026forgetting}. In effect, multi-stage training selects a point on an acquisition--retention trade-off rather than producing a model that is simultaneously strong across domains.

A natural alternative is to train all domains jointly from a stationary mixture, replacing stage ordering with a data-mixture ratio~\citep{liu2025rethinkingmixture,chen2026datamixingsurvey}. Prior work has devoted substantial effort to selecting or adapting these sampling proportions~\citep{xie2023doremi,fan2024doge,liu2025regmix,ye2025mixinglaws}. By keeping every domain active throughout training, joint mixtures avoid the abrupt distribution shifts introduced by stage boundaries---yet they do not automatically recover the strongest outcome of each specialized stage. Different ratios favor different domains, so a direct mixture often settles at an interior compromise (see~\ref{fig:teaser-pareto}). Recent work further shows that domain learning can exhibit threshold behavior as mixture shares vary, while repeated exposure yields scale-dependent returns~\citep{gu2025data,li2026capacityaware,li2026scalingdomaindatarepetition}. Optimizing the mixture ratio therefore searches for a better point on the trade-off rather than eliminating it.

Our work starts with the identification of a central limitation of ratio tuning: a single control variable is asked to serve two fundamentally different roles. Increasing a domain's sampling share changes both its \emph{exposure}---how much data the model sees---and, under the standard unweighted objective, its \emph{optimization scale}---how strongly prediction errors in that domain shape parameter updates. These two quantities are conceptually distinct~\citep{salmani2025samplingloss}, and their preferred allocations need not coincide: the amount of data needed for sufficient coverage may differ from the relative loss weight that yields the strongest joint model. This observation motivates a different intervention: hold the sampled-data mixture fixed and adjust domain loss coefficients directly, decoupling \emph{what the model sees} from \emph{how strongly each domain drives learning}. Figure~\ref{fig:remix-reweight} visualizes this distinction: remixing and reweighting trace different paths through the capability plane.

Figure~\ref{fig:teaser-pareto} illustrates the opportunity created by this decoupling. With the sampled-data mixture and token budget held fixed, a suitable mathematics loss coefficient improves both MMLU and GSM8K over direct mixing---and even surpasses the best value attained on each benchmark along the multi-stage trajectories. A fixed coefficient, however, merely trades one search problem for another: its optimal value is unknown a priori, expensive to sweep at pretraining scale, and may not transfer across data compositions or training configurations. 

We therefore introduce \Method{} (\emph{Loss-Adaptive Reweighting under fixed Exposure}), an online domain-level controller that derives relative loss weights from running domain-loss statistics. After learning-rate warmup, it applies a normalized inverse-square-root rule that adapts relative optimization scale without globally rescaling the training objective. Under this rule, the 7:1 adaptive run reaches a competitive region of the fixed-weight trade-off in a single training run, eliminating the need for a post-hoc oracle grid. The core contributions of this work are threefold: 
\begin{itemize}[leftmargin=10pt,topsep=0em, itemsep=0em]
  \item We disentangle domain exposure from optimization scale in multi-domain pretraining. Under matched 30B-token exposure, stage scheduling, direct mixing, and loss reweighting yield markedly different endpoints; a suitable fixed loss coefficient simultaneously exceeds the best knowledge and mathematical-reasoning scores attained along either stage trajectory (Figure~\ref{fig:teaser-pareto} and Section~\ref{sec:phenomenon}).

  \item We introduce \Method{}, a lightweight online controller that derives domain coefficients from running loss statistics. Its normalized inverse-square-root rule adapts relative optimization scale after warmup while keeping sampled exposure fixed and the mixture-averaged coefficient at one. The controller reaches a competitive region of the fixed-weight trade-off in a single run, without auxiliary search runs (Section~\ref{sec:theory}).

  \item We validate the method across optimizers, sampled-data ratios, domain compositions, and training budgets (Section~\ref{sec:experiments}). At $4\times$ Chinchilla with K:M $=7{:}1$, \Method{} improves MMLU by 0.39 points, GSM8K by 13.57 points, and MATH by 8.40 points. Extending training with mathematics and code to $8\times$ Chinchilla yields gains of 0.83, 17.06, 9.00, 6.71, and 5.60 points on MMLU, GSM8K, MATH, HumanEval, and MBPP, respectively, over direct mixing (Table~\ref{tab:main-results}).
\end{itemize}

%% file: sections/02_phenomenon/section.tex
\section{Mixture Ratios Couple Exposure and Scale}
\label{sec:phenomenon}

Data-mixture ratios determine both which domains the model observes and how strongly their losses enter a standard unweighted update. We first show empirically that changing loss weights at fixed exposure reaches a broader capability frontier than changing the mixture itself. We then formalize why these interventions are not interchangeable and return to matched-exposure training trajectories to isolate their consequences.

\subsection{Reweighting Expands the Remix Frontier}
\label{sec:remix-reweight-frontier}
\begin{figure*}[t]
  \centering
  \includegraphics[width=0.7\textwidth]{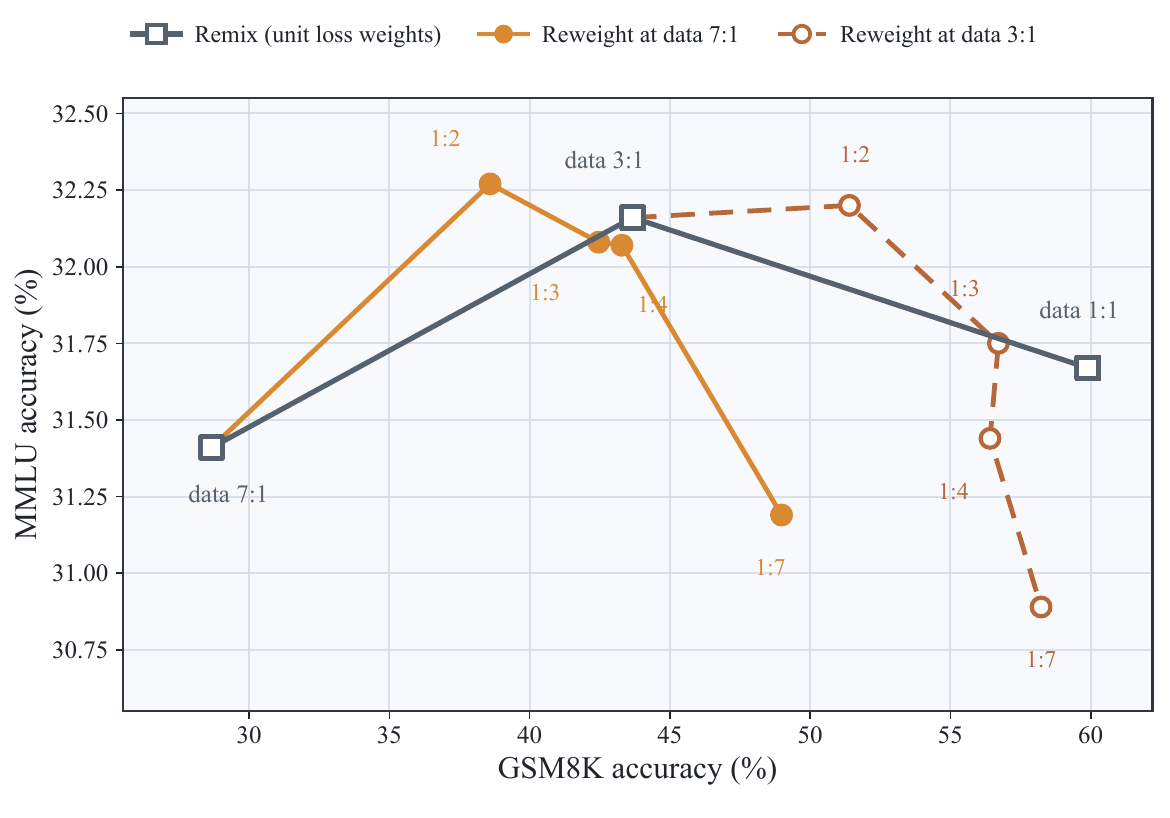}
  \caption{\small\textbf{At fixed exposure, reweighting expands the observed remix frontier.} Gray squares trace direct mixtures with unit loss coefficients at K:M ratios $7{:}1$, $3{:}1$, and $1{:}1$. Warm curves instead fix exposure at $7{:}1$ or $3{:}1$ and vary only the post-warmup loss ratio from $1{:}1$ to $1{:}7$. At both fixed mixtures, $1{:}2$ reweighting improves MMLU and GSM8K simultaneously; the reasoning-heavy $1{:}1$ remix remains the boundary of the measured comparison.
  }
  \label{fig:remix-reweight}
  \vspace{-0.2cm}
\end{figure*}
At a common parameter $\theta_t$, let $m_i$ be the sampling proportion of domain $i$ in a mixed batch of size $B$, and let $\hat g_{i,t}^{(n)}$ be its mean unweighted gradient from $n$ samples. With a shared scalar learning rate $\bar{\eta}$, the two SGD updates are
\begin{equation}
  \begin{aligned}
    \Delta\theta_{t,\mathrm{remix}}
    &=-\bar{\eta}\sum_i m_i'\hat g_{i,t}^{(Bm_i')},\\
    \Delta\theta_{t,\mathrm{reweight}}
    &=-\bar{\eta}\sum_i m_i w_i\hat g_{i,t}^{(Bm_i)}
     =-\sum_i m_i\eta_i\hat g_{i,t}^{(Bm_i)},
    \qquad \eta_i=\bar{\eta}w_i.
  \end{aligned}
  \label{eq:remix-reweight-update}
\end{equation}
Remixing changes both $m_i$ and the number of samples used to estimate its gradient; reweighting keeps the sample allocation fixed and scales the estimate by $w_i$. At fixed $\theta_t$, the expectation of $\hat g_{i,t}^{(n)}$ is the same domain gradient for any $n$, but its variance depends on $n$. Both interventions can change later domain gradients by producing different parameter trajectories. The effective-rate identity is exact for SGD; with nonlinear gradient transformations, it is an abstraction rather than an optimizer-level equality.

Figure~\ref{fig:remix-reweight} reports final-checkpoint MMLU and GSM8K after 30B sampled tokens. The plotted K:M ratios specify full-token exposure before loss masking. Runs share the 1.5B model, learning-rate schedule, and optimizer assignment (Muon for hidden matrices; AdamW for embeddings and output weights). The gray direct-mixture path uses K:M ratios of $7{:}1$, $3{:}1$, and $1{:}1$ at $w_{\mathrm K}=w_{\mathrm M}=1$. The reweighting paths fix K:M at $7{:}1$ or $3{:}1$ and set $w_{\mathrm M}/w_{\mathrm K}\in\{1,2,3,4,7\}$ after the shared 20{,}000-step warmup. Within each ratio, runs use the same frozen training stream and loss masks.

The reweighting curves do not merely add nearby operating points to the remix trade-off. Together they form a broader Pareto envelope across the central operating region: the 7:1 and 3:1 direct anchors are each Pareto-dominated by their corresponding 1:2 reweighted run, and larger coefficients continue to increase mathematical reasoning before eventually trading away knowledge accuracy. The central result is therefore a strict outward expansion of the observed frontier. This separation shows that the sampled-data ratio does not uniquely determine the attainable trade-off and motivates treating exposure and optimization scale as distinct controls.

\subsection{Theory: Why Reweighting Can Outperform Remixing}
\label{sec:exposure-scale}

Equation~\ref{eq:remix-reweight-update} shows the additional degree of freedom available at fixed exposure. We now ask whether changing the mixture under a shared rate can match the loss trade-offs reached by separate domain rates.

\input{sections/02_phenomenon/theory_v3}

\subsection{Same Domain Exposure, Different Outcomes}
\label{sec:same-exposure-outcomes}
\begin{figure*}[t]
  \centering
  \includegraphics[width=0.9\textwidth]{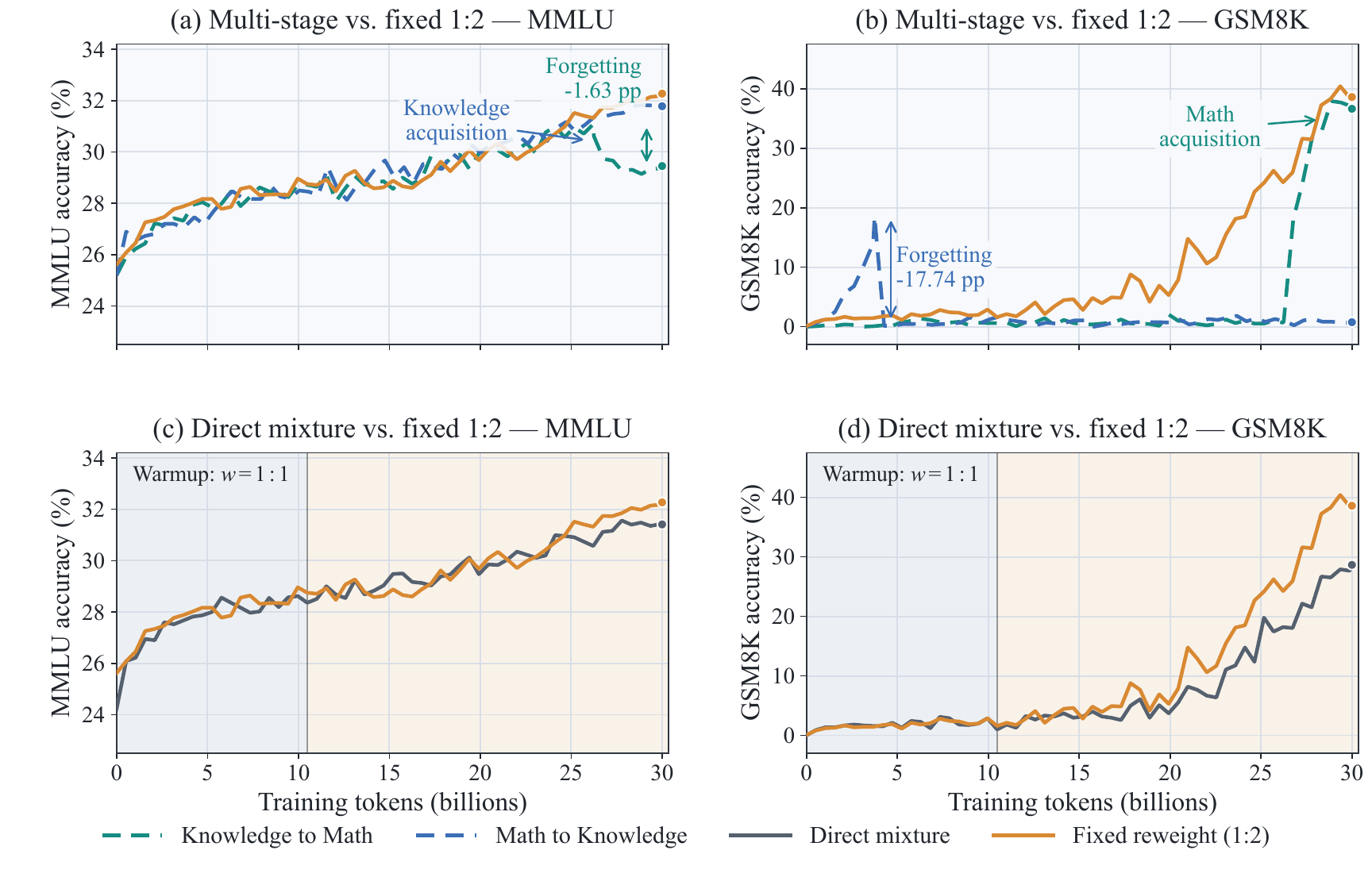}
  \caption{\small\textbf{At matched exposure, stage order determines retention, whereas reweighting improves both capabilities.} Every run receives 26.25B knowledge and 3.75B mathematics tokens. The top row compares stage schedules with fixed reweighting; the bottom row isolates direct mixing versus fixed reweighting. Reversing stage order reverses forgetting. Under the stationary $7{:}1$ mixture, changing the loss ratio from $1{:}1$ to $1{:}2$ moves the final MMLU/GSM8K pair from 31.40/27.67 to 32.27/38.59, above the coordinate-wise maxima of both stage trajectories.
  }
  \label{fig:phenomenon-curves}
\end{figure*}

To test whether cumulative domain exposure alone determines the knowledge--reasoning trade-off, we compare stage scheduling, direct mixing, and loss reweighting under matched token budgets. All runs share the same model configuration, 30B-token budget, learning-rate schedule, and optimizer assignment: Muon for matrix-valued hidden parameters and AdamW for embedding and output parameters. Each run receives 26.25B knowledge tokens and 3.75B mathematics tokens. The two stage baselines present this allocation in opposite orders; the stationary baselines sample both domains throughout training, with direct mixing and fixed reweighting differing only in their relative loss coefficients. Figure~\ref{fig:phenomenon-curves} shows the resulting unsmoothed MMLU and GSM8K checkpoint trajectories; the exact token schedules, final scores, and first-domain drops are reported in Table~\ref{tab:phenomenon} of Appendix~\ref{app:phenomenon-results}. Scalability under different optimizers and training budgets is examined in Section~\ref{sec:experiments}.

\paragraph{Reversing stage order reverses the trade-off.} Knowledge $\rightarrow$ Math ends at 29.45 MMLU and 36.62 GSM8K after a 1.63-point MMLU drop; Math $\rightarrow$ Knowledge ends at 31.78 and 0.76 after a 17.74-point GSM8K drop. In each case, the second stage erodes performance on the first domain.

\paragraph{Direct mixing moderates rather than resolves the trade-off.} The stationary $7{:}1$ mixture ends at 31.40 MMLU and 27.67 GSM8K, avoiding the severe GSM8K collapse of Math $\rightarrow$ Knowledge. Still, both scores remain below the respective maxima reached along the two multi-stage trajectories.

\paragraph{Loss reweighting improves both capabilities at fixed exposure.} Among the fixed coefficients in Figure~\ref{fig:teaser-pareto}, $w_{\mathrm{K}}:w_{\mathrm{M}}=1{:}2$ attains the highest MMLU and is used here. At the same $7{:}1$ exposure, it reaches 32.27 MMLU and 38.59 GSM8K, improving on direct mixing by 0.87 and 10.92 points, respectively, and exceeding the best scores attained anywhere along the two stage trajectories (31.85 and 37.91).

Together, these matched-exposure runs show that domain token counts alone do not determine the knowledge--reasoning outcome: stage order changes which capability is retained, while direct mixing yields a compromise. Reweighting the same stationary mixture improves both metrics, identifying domain loss scale as a separate control over the trade-off.

%% file: sections/02_phenomenon/theory_v3.tex
\begin{figure*}[t]
  \centering
  \includegraphics[width=0.4\textwidth]{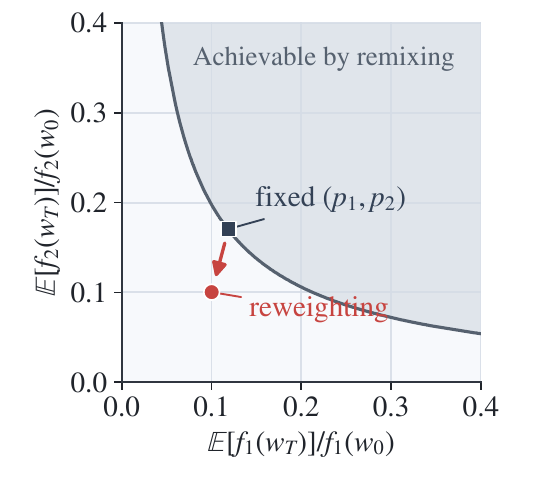}
  \includegraphics[width=0.4\textwidth]{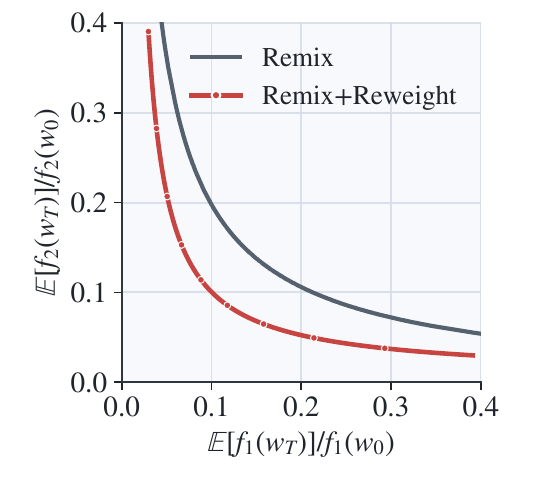}
  \caption{\small\textbf{Remixing cannot subsume reweighting.} Expected losses for the two-domain linear regression model, with \(B=128\), \(T=20\), block dimensions \(d_1=d_2=512\), and expected curvatures of $2.5$ and $1$ for the two domains, respectively. \textbf{Left:} At fixed \(p_1=p_2=1/2\), reweighting improves both domain losses and is not Pareto-dominated by any choice of mixture and shared learning rate. \textbf{Right:} Reweighting improves the Pareto frontier achievable by remixing alone.}
  \label{fig:quadratic-pareto-separation}
\end{figure*}
We formalize this distinction in a two-domain linear regression model. Let \(A_1,A_2\in\R^{d\times d}\) be nonzero symmetric positive semidefinite matrices with orthogonal supports, i.e., \(A_1A_2=0\). In domain \(j\in\{1,2\}\), samples follow \(x_j\sim\mathcal N(0,A_j)\) and \(y_j=x_j^\top w^\star\). Using a common \(w^\star\) is without loss of generality: if the two domains have different ground-truth parameters, their projections onto the respective orthogonal supports can be summed to obtain a shared parameter vector. We consider the MSE loss, and the population loss in domain \(j\) is
\[
  f_j(w)=\tfrac12\,\E_{x_j}\!\bigl[(x_j^\top w-y_j)^2\bigr]
        =\tfrac12\,(w-w^\star)^\top A_j(w-w^\star).
\]
Starting from $w_0$ with $f_1(w_0),f_2(w_0)>0$, we run $T\geq1$ SGD steps with batches of $B\geq2$ samples. Domain $j$ supplies $Bp_j\in\mathbb N^+$ samples per batch, where $p_1+p_2=1$ and $p_j\in(0,1)$. Write $L_j(p_j,\eta_j)=\E[f_j(w_T)]$ for its expected terminal loss under domain rate $\eta_j$. Orthogonal supports make this loss depend only on $p_j$ and $\eta_j$; the full update and sampling assumptions are in Appendix~\ref{app:proof-v3-main}.

We compare the following two strategies in this simplified setup:
\begin{itemize}
    \item \textbf{Remixing} varies mixture proportions \(p_1,p_2\in(0,1)\) while applying a shared learning rate $\eta$ to both domains, which may be tuned for each choice of mixture proportions. 
    \item \textbf{Reweighting} instead keeps the mixture proportions fixed and tunes learning rates $\eta_1$ and $\eta_2$ separately for the two domains.
\end{itemize}




\begin{proposition}[Remixing cannot dominate reweighting]
\label{prop:v3-main}
Fix any feasible mixture \(p_1, p_2\). Let $ \eta_1^\star\in\argmin_{\eta_1} L_1(p_1,\eta_1),
 \eta_2^\star\in\argmin_{\eta_2} L_2(p_2,\eta_2)$. 
 Then the optimal domain losses achieved by reweighting $
\bigl(L_1(p_1,\eta_1^\star),
L_2(p_2,\eta_2^\star)
\bigr)$
cannot be Pareto-dominated by remixing.
Specifically, for every feasible mixture \((p_1',p_2')\not = (p_1,p_2)\) and every shared step size \(\eta'\),
\begin{equation}
\label{eq:v3-remix-separation}
L_1(p_1',\eta')
>
L_1(p_1,\eta_1^\star)
\quad\text{or}\quad
L_2(p_2',\eta')
>
L_2(p_2,\eta_2^\star).
\end{equation}
\end{proposition}

The proof is provided in Appendix~\ref{app:proof-v3-main}. 

Figure~\ref{fig:quadratic-pareto-separation} illustrates Proposition~\ref{prop:v3-main} with a concrete instance of the model.  The eigenvalues of \(A_1\) and \(A_2\) are drawn i.i.d.\ and uniformly over \([1.25,\,3.75]\) and \([0.5,\,1.5]\), respectively. At fixed mixture \(p_1=p_2=1/2\), reweighting improves both domain losses, and no choice of mixture and shared learning rate can simultaneously match both losses. Applying reweighting improves the Pareto frontier achievable by remixing. Details are providedd in Appendix~\ref{app:v3-numerics}. 

%% file: sections/03_theory_method/section.tex
\section{\Method{}: Loss-Adaptive Reweighting under Fixed Exposure}
\label{sec:theory}

Section~\ref{sec:phenomenon} establishes that loss scale matters, but a fixed weight does not provide a practical rule for choosing the coefficients. Grid search requires a separate pretraining run for each candidate, and the best setting may not transfer across data mixtures, compute budgets, or optimizers. We therefore seek an online alternative that infers and adapts relative domain weights from signals already available during training, eliminating auxiliary search runs.

\subsection{From Domain Loss to Optimization Scale}
\label{sec:toy-invsqrt}

We derive the inverse-square-root exponent from stochastic-gradient noise near a domain optimum. Consider binary logistic regression with $x\sim\mathcal N(0,I_d)$, true label $y_0=\operatorname{sign}(u^\top x)$ for a unit vector $u$, and observed label $y$ obtained by independently flipping $y_0$ with probability $\rho\in(0,1/2)$. Let $r_\rho^\star$ minimize the population loss along $w=ru$ over $r\geq0$, write $w_\rho^\star=r_\rho^\star u$ and $\ell_\rho^\star=\E\log(1+e^{-y(w_\rho^\star)^\top x})$, and let $\Sigma_\rho$ be the per-example gradient covariance at $w_\rho^\star$.

\begin{theorem}[Loss--gradient-noise scaling]
\label{thm:loss-noise-scaling}
For the Gaussian model above with $d\geq2$, as $\rho\downarrow0$,
\begin{equation}
  \ell_\rho^\star=\Theta(\sqrt{\rho}),\qquad
  u^\top\Sigma_\rho u=\Theta\!\left((\ell_\rho^\star)^2\right),\qquad
  v^\top\Sigma_\rho v=\Theta(\ell_\rho^\star)
  \label{eq:loss-noise-scaling}
\end{equation}
for every unit vector $v\perp u$. Consequently, $\operatorname{tr}(\Sigma_\rho)=\Theta((d-1)\ell_\rho^\star+(\ell_\rho^\star)^2)$, so the root-mean-square stochastic-gradient norm scales as $\Theta(\sqrt{d\ell_\rho^\star})$. Appendix~\ref{app:proof-loss-noise-scaling} gives the complete derivation and constants.
\end{theorem}

Because a loss coefficient $w_i$ scales the root-mean-square gradient linearly, Theorem~\ref{thm:loss-noise-scaling} suggests, for comparable effective dimensions and domain constants,
\begin{equation}
  \operatorname{RMS}(w_i g_i)
  =\Theta\!\left(w_i\sqrt{d_i\ell_i^\star}\right),
  \qquad
  w_i\propto(\ell_i^\star)^{-1/2},
  \label{eq:inverse-sqrt-motivation}
\end{equation}
which offsets the larger noise scale of a high-loss domain. This stylized model motivates the exponent rather than an exact language-model scaling law. Since $\ell_i^\star$ is unavailable and domain difficulty changes during training, \Method{} uses smoothed current losses and normalizes the resulting coefficients as described next.

\subsection{The Adaptive Reweighting Rule}
\label{sec:method}

At step $t$, \Method{} tracks exponential moving averages $\hat{\ell}_{i,t}$ of the mean unweighted cross-entropy on unmasked domain-$i$ targets and $\hat{c}_{i,t}$ of their counts. The estimated loss-token share is $\hat{\pi}_{i,t}=\hat{c}_{i,t}/\sum_j\hat{c}_{j,t}$. All coefficients remain one during learning-rate warmup; afterward, the next-step weights are
\begin{equation}
  r_{i,t}=(\hat{\ell}_{i,t}+\epsilon)^{-1/2},
  \qquad
  w_{i,t+1}=\frac{r_{i,t}}{\sum_j \hat{\pi}_{j,t}r_{j,t}} .
\label{eq:selected-rule}
\end{equation}
Here $\epsilon>0$ ensures numerical stability. The normalization keeps $\sum_i\hat{\pi}_{i,t}w_{i,t+1}=1$, changing the relative domain loss scales without changing their mixture-averaged coefficient. Algorithm~\ref{alg:online-invsqrt} gives the full update, including absent-domain handling; Appendix~\ref{app:hyperparameters} reports the settings.

\subsection{Validating the Reweighting Dynamics}

\begin{figure*}[t]
\centering
\includegraphics[width=0.9\textwidth]{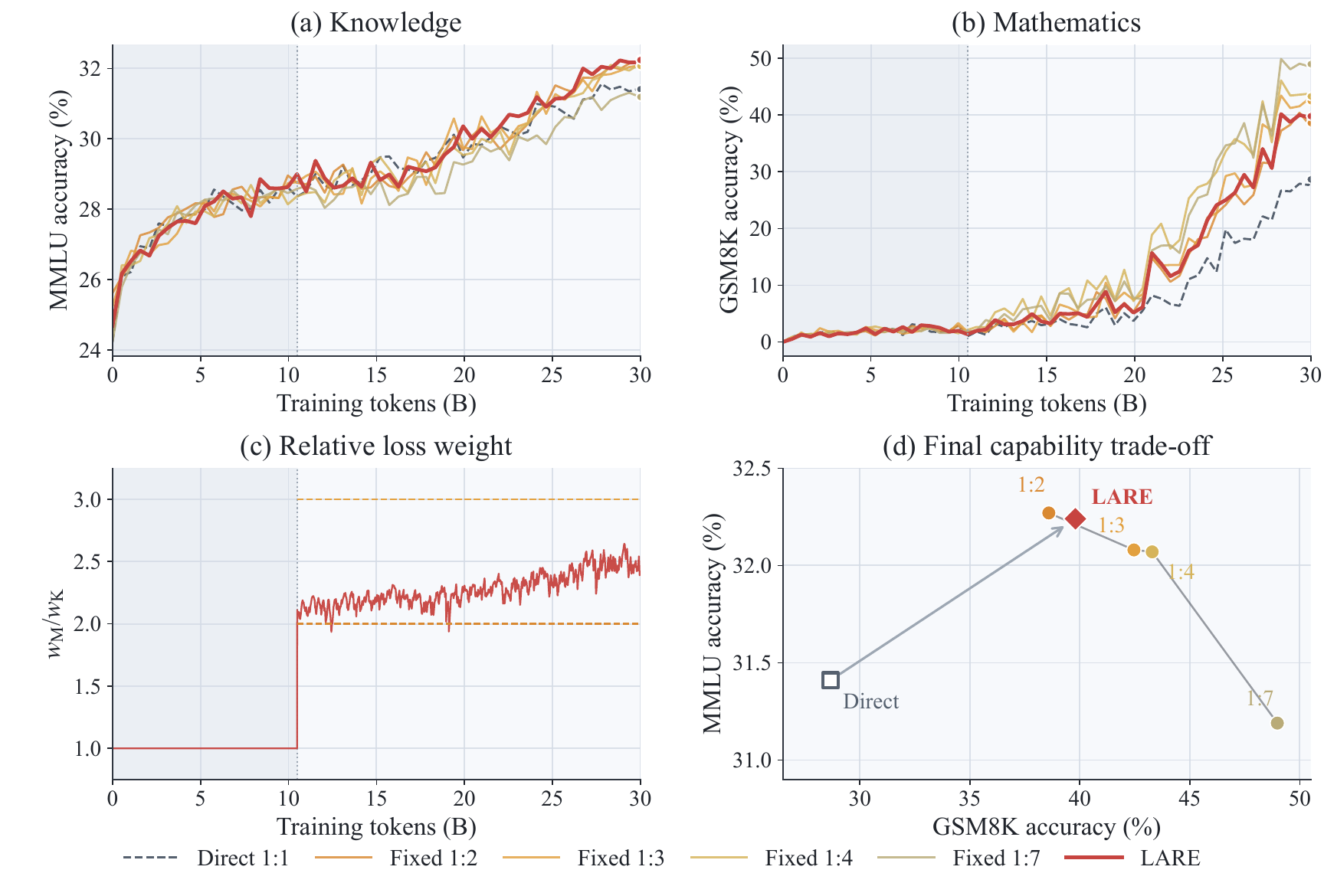}
\caption{\small\textbf{\Method{} selects a stable, competitive loss scale in one run.} The top row shows unsmoothed MMLU and GSM8K trajectories; the bottom row shows the selected relative weight and final endpoints. After the shared 20,000-step warmup at $w_{\mathrm M}/w_{\mathrm K}=1$, the ratio averages 2.30 from step~25{,}000 onward and ends at 2.41, between the fixed $1{:}2$ and $1{:}3$ references. \Method{} moves the direct endpoint from 31.40/27.67 to 32.24/39.80 without an auxiliary coefficient sweep.}
\label{fig:loss-scale-dynamics}
\vspace{-0.4cm}
\end{figure*}

We validate \Method{} in the controlled 30B-token, 7:1 knowledge--mathematics setting used for the fixed-weight sweep. The model, sampling distribution, optimizer, learning-rate schedule, and token budget are all held constant; the loss-weighting rule is the only experimental variable. Figure~\ref{fig:loss-scale-dynamics} compares MMLU and GSM8K trajectories of the full fixed-weight grid against \Method{}, alongside the relative weight $w_{\mathrm M,t}/w_{\mathrm K,t}$ selected online. Together, the three panels address two questions: whether the controller is stable after warmup, and whether the scale it selects occupies a useful region of the fixed-weight trade-off.

After warmup, the selected weight ratio remains stable near the fixed $1{:}2$ and $1{:}3$ settings (Figure~\ref{fig:loss-scale-dynamics}). At the final checkpoint, \Method{} improves direct $1{:}1$ weighting by 0.84 MMLU and 12.13 GSM8K points. Its endpoint is competitive with fixed $1{:}2$, showing that the controller can select a useful loss scale in one run without an auxiliary sweep.


%% file: sections/04_experiments/section.tex
\section{Experiments: Ablations and Scaling up}
\label{sec:experiments}

\subsection{Experimental Setup}
\label{sec:setup}

All experiments use a common 1.5B GPT-2-style decoder-only Transformer~\citep{radford2019language} and organize the training corpus into knowledge (K), mathematics (M), and code (C). Mixture ratios denote full tokens sampled before loss masking, while $1\times$, $4\times$, and $8\times$ Chinchilla~\citep{hoffmann2022training} correspond to 30B, 120B, and 240B sampled tokens. Following the controlled optimizer comparison in Section~\ref{sec:optimizer-ablation}, all subsequent main experiments use \agd{}. Unless stated otherwise, each reported cell corresponds to a single run evaluated at its final checkpoint. Appendix~\ref{app:data} specifies data construction and masking; Appendices~\ref{app:details} and~\ref{app:agd} provide detailed model, training, and optimizer settings, while Appendix~\ref{app:evaluation-metrics} specifies evaluation protocols.

\subsection{Ablating over Optimizers}
\label{sec:optimizer-ablation}

\begin{algorithm}[t]
\caption{Row-wise $(2,\infty)$ \agd{} update for a linear weight matrix.}
\label{alg:actgd}
\begin{algorithmic}[1]
\Require Weight $W\in\mathbb{R}^{d_{\mathrm{out}}\times d_{\mathrm{in}}}$; batch gradient $G_t$; uncentered input second moment $C_t$; EMA states $H_{t-1},S_{t-1}$ initialized at zero; decays $\beta_g,\beta_c$; learning rate $\eta_t$; weight decay $\lambda$; numerical constant $\epsilon$
\State $H_t\gets\beta_gH_{t-1}+(1-\beta_g)G_t$; $S_t\gets\beta_cS_{t-1}+(1-\beta_c)C_t$
\State $\widehat H_t\gets H_t/(1-\beta_g^t)$; $\widehat S_t\gets S_t/(1-\beta_c^t)$
\State $A_t\gets\Call{NewtonSchulzInverse}{\widehat S_t,\epsilon}$
\State $U_t\gets\widehat H_tA_t$
\For{$j=1,\ldots,d_{\mathrm{out}}$}
  \State $D_{t,j}\gets U_{t,j}/\bigl(\sqrt{\max\{0,\widehat H_{t,j}U_{t,j}^\top\}}+\epsilon\bigr)$
\EndFor
\State $W\gets(1-\eta_t\lambda)W-\eta_tD_t$
\end{algorithmic}
\end{algorithm}

\begin{table}[t]
  \caption{\small\textbf{ActGD gives the strongest direct-mixture result while using a single optimizer-specific hyperparameter set.} All runs use the same 1.5B model, $1\times$ Chinchilla budget, and K:M $=7{:}1$ mixture. ``Blocks'' and ``Embed./head'' identify the update applied to transformer-block matrices and vocabulary-facing weights, respectively. A checkmark indicates that both groups share one optimizer family and hyperparameter set; a cross indicates a hybrid configuration with separately tuned settings. Bold denotes the best benchmark result.}
  \label{tab:optimizer-ablation}
  \begin{center}
  \resizebox{0.7\columnwidth}{!}{%
  \begin{tabular}{lcccrr}
    \toprule
    Optimizer & Blocks & Embed./head & \shortstack{Single HP set} & MMLU & GSM8K \\
    \midrule
    AdamW & AdamW & AdamW & $\checkmark$ & 31.30 & 21.91 \\
    Muon & Muon & AdamW & $\times$ & 31.40 & 27.67 \\
    \agd{} & \agd{} & \agd{} & $\checkmark$ & \textbf{31.82} & \textbf{28.58} \\
    \bottomrule
  \end{tabular}}
  \end{center}
\end{table}

Alongside \Method{}, we introduce Activation Gradient Descent (\agd{}), a base optimizer that constructs row-wise updates in a whitened activation geometry and applies the same update family to every learned matrix. Algorithm~\ref{alg:actgd} summarizes the practical update, while Appendix~\ref{app:agd} provides its derivation and implementation details.

To select the optimizer for the subsequent experiments, we compare AdamW~\citep{loshchilov2019decoupled}, Muon~\citep{liu2025muon}, and \agd{} under the same 1.5B model, $1\times$ Chinchilla budget, and K:M $=7{:}1$ direct-mixture objective. This comparison deliberately excludes \Method{} so that it isolates the base optimizer rather than its interaction with loss reweighting. Because optimal hyperparameters do not generally transfer across optimizers, unequal tuning can confound the comparison~\citep{wen2025fantasticpretrainingoptimizers}. We therefore sweep the learning rate or rates independently for each optimizer and select its best configuration at the full training budget instead of transferring a common learning rate; for the hybrid Muon setup, the Muon and auxiliary AdamW learning rates are tuned separately. All remaining model, data, and schedule settings are matched.

Table~\ref{tab:optimizer-ablation} shows that \agd{} obtains the highest MMLU and GSM8K scores, outperforming Muon by 0.42 and 0.91 points and AdamW by 0.52 and 6.67 points, respectively. Beyond this accuracy advantage, \agd{} applies the same activation-aware update family to transformer blocks, token embeddings, and the output head, using a single shared hyperparameter set throughout. Muon instead combines a Muon block update with an auxiliary AdamW optimizer for the vocabulary-facing parameters, requiring the two components to be tuned separately. The unified ActGD configuration therefore reduces the optimizer search space and simplifies hyperparameter tuning. Accordingly, all subsequent main-text experiments use \agd{}; results with the other optimizers are reported in Appendix~\ref{app:optimizer-details}.

\begin{table*}[t]
  \caption{\small\textbf{\Method{} improves all matched $4\times$ and $8\times$ benchmarks.} Mean gains over direct mixing are 7.45 points for K:M at $4\times$ Chinchilla and 5.90 and 7.84 points for K:M:C at $4\times$ and $8\times$ Chinchilla, respectively. At $8\times$ Chinchilla, our model exceeds Qwen1.5-1.8B, pretrained on 2.2T tokens (over $9\times$ our token budget), by 9.22 points on average across five benchmarks. Direct--\Method{} pairs match model, sampled mixture, optimizer, and token budget. Avg. $\Delta$ averages gains over available benchmarks; bold marks the higher score within each matched pair.}
  \label{tab:main-results}
  \begin{center}
  \resizebox{\textwidth}{!}{%
  \begin{tabular}{llcrcrrrrrr}
    \toprule
    Model & Objective & Size & Tokens & Mixture
      & MMLU & GSM8K & MATH & HumanEval & MBPP & Avg. $\Delta$ (pp) \\
    \midrule
    Ours & Direct mixture & 1.5B & 120B & K:M $=7{:}1$
      & 33.97 & 37.00 & 17.20 & -- & -- & -- \\
    Ours & \Method{} & 1.5B & 120B & K:M $=7{:}1$
      & \textbf{34.36} & \textbf{50.57} & \textbf{25.60} & -- & -- & $+7.45$ \\
    \midrule
    Ours & Direct mixture & 1.5B & 120B & K:M:C $=7{:}0.5{:}0.5$
      & 33.52 & 26.76 & 13.80 & 15.85 & 9.40 & -- \\
    Ours & \Method{} & 1.5B & 120B & K:M:C $=7{:}0.5{:}0.5$
      & \textbf{33.53} & \textbf{41.55} & \textbf{20.20} & \textbf{21.34} & \textbf{12.20} & $+5.90$ \\
    \midrule
    Ours & Direct mixture & 1.5B & 240B & K:M:C $=7{:}0.5{:}0.5$
      & 34.13 & 29.72 & 15.80 & 17.68 & 9.80 & -- \\
    Ours & \Method{} & 1.5B & 240B & K:M:C $=7{:}0.5{:}0.5$
      & \textbf{34.96} & \textbf{46.78} & \textbf{24.80} & \textbf{24.39} & \textbf{15.40} & $+7.84$ \\
    \midrule
    Qwen1.5 & Base & 1.8B & 2.2T & external
      & 33.21 & 31.46 & 9.20 & 23.17 & 3.20 & -- \\
    Qwen1.5 & Base & 7B & 2.4T & external
      & 42.31 & 51.63 & 14.80 & 44.51 & 39.80 & -- \\
    \bottomrule
  \end{tabular}}
  \end{center}
\end{table*}


\subsection{Scaling Across Training Budgets and Domains}
\label{sec:main-results}

\paragraph{Reweighting gains persist at $4\times$ Chinchilla.}
At $4\times$ Chinchilla, \Method{} improves all three reported benchmarks over the matched K:M $=7{:}1$ direct mixture (Table~\ref{tab:main-results}). MMLU rises modestly from 33.97 to 34.36, while GSM8K gains 13.57 points and MATH gains 8.40 points. The reasoning benefit seen at $1\times$ Chinchilla thus persists at the larger budget without a knowledge decline.

\paragraph{\Method{} also improves multiple domains.}
With code added at K:M:C $=7{:}0.5{:}0.5$, \Method{} improves every measured benchmark at both $4\times$ and $8\times$ Chinchilla (Table~\ref{tab:main-results}). The five-benchmark mean gain grows from 5.90 to 7.84 points: the GSM8K gain rises from 14.79 to 17.06 points, the MBPP gain from 2.80 to 5.60 points, and the MMLU gain from 0.01 to 0.83 points. The larger gains at $8\times$ than at $4\times$ on all five benchmarks indicate that \Method{}'s advantage increases over the longer training horizon tested here. Appendix~\ref{app:long-budget-trajectories} shows the MMLU, GSM8K, and MBPP training curves. The Qwen1.5 base models~\citep{qwen2024qwen15} provide external scale references under the same evaluation protocol (Appendix~\ref{app:evaluation-metrics}).

\subsection{Further Analyses}
\label{sec:ratio-ablation}
We also examine data-mixture robustness (Appendix~\ref{app:ratio-results}), reweighting across optimizers and fixed-weight baselines (Appendix~\ref{app:optimizer-details}), and scaling across model sizes at their corresponding Chinchilla budgets (Appendix~\ref{app:scaling}).

%% file: sections/06_conclusion/section.tex
\section{Conclusion}

Multi-domain pretraining has two controls that standard mixture design conflates: domain exposure and optimization scale. At fixed exposure, loss reweighting produces endpoints beyond the observed stage schedules and expands the observed remix Pareto frontier. Our analysis attributes this separation to domain-specific effective learning rates, which can reach loss trade-offs unavailable to a shared-rate mixture. We therefore introduce \Method{}, an online controller that converts running domain losses into normalized inverse-square-root weights while preserving the sampled mixture and fixing the mixture-averaged coefficient at one. \Method{} selects a competitive scale in one run without a coefficient sweep. Experiments cover base optimizers, mixture ratios, domain compositions, and 30B--240B token budgets. At $8\times$ Chinchilla, the 1.5B model improves all five evaluated benchmarks by 7.84 points on average over matched direct mixing, and every gain exceeds its $4\times$ counterpart. These results support treating data allocation and loss scaling as separate pretraining design variables: exposure determines what the model sees, whereas reweighting determines how strongly each domain shapes learning.

%% file: sections/05_related_work/section.tex
\section{Related Work}
\label{sec:related-work}

\paragraph{Multi-stage pretraining.}
Domain-adaptive pretraining and modern midtraining recipes introduce specialized data in later stages to strengthen target capabilities~\citep{gururangan2020dont,walsh2025olmo2,liu2025midtraining,lu2024mathcoder2}. However, this temporal specialization raises a fundamental retention question: studies of continual pretraining report substantial forgetting in some regimes, while showing that self-supervised continuation can preserve prior knowledge more robustly than supervised adaptation in others~\citep{ke2023continualpretraininglanguagemodels,yildiz2025continual,cossu2022continual,marek2026forgetting}. These results indicate that forgetting is regime-dependent rather than an inevitable consequence of continued training.

\paragraph{Data-mixture ratio selection.}
The dominant approach to multi-domain data mixing optimizes the sampling ratio itself~\citep{liu2025rethinkingmixture,chen2026datamixingsurvey}. Offline methods such as DoReMi and DoGE infer domain proportions from proxy losses or generalization estimates, while MDE evaluates candidate mixtures with domain-expert models~\citep{xie2023doremi,fan2024doge,belenki2025mde}. Online methods instead adapt sampling during training: Skill-it! builds skill curricula, DGA uses target-gradient alignment, Data Mixing Agent learns a transferable policy, and AC-ODM and HDS optimize schedules from domain interactions or multiple training signals~\citep{chen2023skillit,fan2024dga,yang2026datamixingagent,ma2026acodm,dang2026hds}. Further work refines the control space by reconstructing domains from topics, estimating importance in a representation space, modeling domain interactions, or unifying multiple data operations~\citep{peng2025topic,xie2025chameleon,xu2026dograph,liang2026dataflex}. All of these works change \emph{what} the model sees. Loss reweighting offers a complementary lever: it changes the contribution of already-sampled data to the gradient, with precedents in task-level balancing and example-level language-model training~\citep{kendall2018multitask,chen2018gradnorm,sener2018multitask,sow2025dynamic}. Importantly, sampling and loss coefficients can interact through gradient variance and generalization, and may become coupled under importance correction or particular bilevel formulations~\citep{salmani2025samplingloss,needell2014weighted,tan2026fastmix}.

\paragraph{Scaling laws for data mixtures.}
A ratio selected in one configuration need not remain optimal as model size, training budget, or domain reuse changes. RegMix and Data Mixing Laws fit performance-response models from smaller proxy runs, and recent studies explicitly characterize how useful allocations shift with scale, model capacity, and repeated use of limited domain data~\citep{liu2025regmix,ye2025mixinglaws,gu2025data,kang2025autoscale,li2026capacityaware,li2026scalingdomaindatarepetition}. Related systems seek lower-cost transfer across configurations: Olmix reuses mixture knowledge as the available domain set evolves, whereas DeMix and MergeMix use separately trained domain models or model merging as proxies for mixture quality~\citep{mchen2026olmix,li2026demix,wang2026mergemix}. This body of work underscores why a single sampling ratio is difficult to transfer across scales.

%% file: sections/90_appendix_theory/section.tex

\section{Theory}
\label{app:theory}

\subsection{Proof of the Loss--Gradient-Noise Scaling}
\label{app:proof-loss-noise-scaling}

We prove Theorem~\ref{thm:loss-noise-scaling} and make explicit the assumptions behind its use as motivation for LARE. Let $x\sim\mathcal N(0,I_d)$, fix a unit vector $u$ that defines the ground-truth decision boundary, and set the ground-truth label to $y_0=\operatorname{sign}(u^\top x)$. Independently flip $y_0$ with probability $\rho\in(0,1/2)$ to obtain the observed label $y$. We optimize the scale of the logistic classifier $w=ru$, $r\geq0$, and write $Z=u^\top x\sim\mathcal N(0,1)$ and $T=|Z|$. The population logistic risk is exactly
\begin{equation}
  L_\rho(r)
  =(1-\rho)\E\log(1+e^{-rT})
   +\rho\E\log(1+e^{rT})
  =\E\log(1+e^{-rT})+\rho r\E T.
  \label{eq:gaussian-logistic-risk}
\end{equation}
The second equality uses $\log(1+e^a)=a+\log(1+e^{-a})$. The risk is strictly convex in $r$ and, for $\rho<1/2$, its unique minimizer $r_\rho^\star$ satisfies
\begin{equation}
  \E\!\left[\frac{T}{1+e^{r_\rho^\star T}}\right]
  =\rho\E T.
  \label{eq:gaussian-logistic-stationarity}
\end{equation}
As $\rho\downarrow0$, this identity forces $r_\rho^\star\to\infty$. With $\phi$ denoting the standard-normal density, the substitution $s=rT$ and dominated convergence give
\begin{align}
  \E\!\left[\frac{T}{1+e^{rT}}\right]
  &=\frac{2}{r^2}\int_0^\infty\frac{s\phi(s/r)}{1+e^s}\,ds
    =\frac{\pi^2}{6\sqrt{2\pi}}\frac{1+o(1)}{r^2},
  \label{eq:stationarity-asymptotic}\\
  \E\log(1+e^{-rT})
  &=\frac{2}{r}\int_0^\infty\phi(s/r)\log(1+e^{-s})\,ds
    =\frac{\pi^2}{6\sqrt{2\pi}}\frac{1+o(1)}{r}.
  \label{eq:clean-risk-asymptotic}
\end{align}
Here we used $\int_0^\infty s/(1+e^s)\,ds=\int_0^\infty\log(1+e^{-s})\,ds=\pi^2/12$ and $\E T=\sqrt{2/\pi}$. Substituting Equation~\ref{eq:stationarity-asymptotic} into Equation~\ref{eq:gaussian-logistic-stationarity}, and then using Equation~\ref{eq:clean-risk-asymptotic} in Equation~\ref{eq:gaussian-logistic-risk}, yields
\begin{equation}
  r_\rho^\star=\frac{\pi}{\sqrt{12\rho}}\bigl(1+o(1)\bigr),
  \qquad
  \ell_\rho^\star=L_\rho(r_\rho^\star)
  =\sqrt{\frac{2\pi}{3}}\sqrt{\rho}+o(\sqrt{\rho}).
  \label{eq:optimal-scale-and-loss}
\end{equation}

We next compute the covariance of the stochastic gradient with respect to the full parameter vector $w\in\mathbb R^d$. For one example,
\begin{equation}
  g(w;x,y)=\nabla_w\log(1+e^{-yw^\top x})
  =-\frac{yx}{1+e^{yw^\top x}}.
  \label{eq:logistic-example-gradient}
\end{equation}
Decompose $x=Zu+x_\perp$, where $x_\perp\sim\mathcal N(0,I_d-uu^\top)$ is independent of $Z$ and of the label flip. At $w=ru$, the scalar multiplying $x$ in the gradient depends only on $Z$ and the label flip, so the orthogonal component has zero mean. Averaging the parallel component over the label flip gives
\begin{equation}
  \E g(ru;x,y)
  =u\left(\rho\E T-\E\!\left[\frac{T}{1+e^{rT}}\right]\right).
  \label{eq:gaussian-logistic-mean-gradient}
\end{equation}
Equation~\ref{eq:gaussian-logistic-stationarity} therefore implies $\E g(w_\rho^\star;x,y)=0$ at $w_\rho^\star=r_\rho^\star u$. By convexity of the population logistic risk in $w$, this point is also a minimizer over $\mathbb R^d$. Its gradient covariance thus equals its second moment. Conditional on $T$, the label-averaged squared scalar multiplying $xx^\top$ at $w=ru$ is
\begin{equation}
  a_{\rho,r}(T)
  =\frac{1-\rho}{(1+e^{rT})^2}
   +\frac{\rho}{(1+e^{-rT})^2}.
  \label{eq:gradient-noise-scalar}
\end{equation}
Independence and rotational symmetry then give
\begin{equation}
  \Sigma_\rho
  =\alpha_{\parallel}uu^\top
   +\alpha_{\perp}(I_d-uu^\top),
  \qquad
  \alpha_{\parallel}=\E[T^2a_{\rho,r_\rho^\star}(T)],
  \quad
  \alpha_{\perp}=\E[a_{\rho,r_\rho^\star}(T)].
  \label{eq:gradient-noise-decomposition}
\end{equation}
For the clean-label term in $\alpha_\perp$, the same substitution $s=rT$ gives
\begin{equation}
  \E\frac{1}{(1+e^{rT})^2}
  =\frac{2\phi(0)}{r}\int_0^\infty\frac{1}{(1+e^s)^2}\,ds+o(r^{-1})
  =\Theta(r^{-1}),
  \label{eq:orthogonal-clean-noise}
\end{equation}
where the integral equals $\log2-1/2>0$. The flipped-label term converges to $\rho$, because $(1+e^{-rT})^{-2}\to1$ almost surely and is bounded by one. Hence $\alpha_\perp=\Theta(r^{-1}+\rho)=\Theta(\sqrt\rho)$. Along the ground-truth direction, the additional factor $T^2$ changes the unflipped-label term to $\Theta(r^{-3})$, whereas dominated convergence makes the flipped-label term asymptotic to $\rho\E T^2=\rho$. Therefore
\begin{equation}
  \alpha_\perp=\Theta(\sqrt\rho)=\Theta(\ell_\rho^\star),
  \qquad
  \alpha_\parallel=\Theta(\rho)=\Theta\!\left((\ell_\rho^\star)^2\right).
  \label{eq:gradient-noise-eigenvalue-scaling}
\end{equation}
Equation~\ref{eq:loss-noise-scaling} follows immediately. Moreover, $\|\Sigma_\rho\|_{\mathrm{op}}=\Theta(\ell_\rho^\star)$ and $\operatorname{tr}(\Sigma_\rho)=(d-1)\alpha_\perp+\alpha_\parallel=\Theta((d-1)\ell_\rho^\star+(\ell_\rho^\star)^2)$. Since the population gradient vanishes at the optimum, $\E\|g(w_\rho^\star;x,y)\|^2=\operatorname{tr}(\Sigma_\rho)$, which establishes the stated root-mean-square scaling. \qed

\paragraph{Scope of the argument.} The low-loss limit relies on optimizing the classifier scale $r$. If $\|w\|$ were fixed to one, even perfectly separable Gaussian halfspace data would retain the strictly positive logistic loss $\E\log(1+e^{-|Z|})$, so no $\ell_\rho^\star\to0$ regime would exist. The theorem is therefore a mechanism result for unconstrained scale optimization, not a claim about unit-normalized classifiers. It also distinguishes the gradient-noise variance, which is linear in loss in the dominant orthogonal subspace, from its standard deviation, which is proportional to the square root of loss. Multiplying a domain loss by $w_i$ scales the latter by $w_i$, motivating $w_i\propto(\ell_i^\star)^{-1/2}$ when the effective dimensions and hidden constants are comparable. LARE replaces the unavailable optimum $\ell_i^\star$ by an online smoothed loss; this substitution is a practical modeling choice rather than a consequence of the asymptotic theorem.

\begin{algorithm}[t]
\caption{\Method{}: online domain-level loss-adaptive reweighting under fixed exposure.}
\label{alg:online-invsqrt}
\begin{algorithmic}[1]
\Require Domains $\mathcal D$; optimizer $\mathcal O$; EMA decay $\beta$; numerical constant $\epsilon$; learning-rate warmup length $T_{\mathrm w}$
\State $\hat{\ell}_{i,0}\gets 0$, $\hat{c}_{i,0}\gets 0$,
  $w_{i,1}\gets 1$, $\mathsf{seen}_i\gets\mathsf{false}$ for all
  $i\in\mathcal D$
\For{optimizer steps $t=1,\ldots,T$}
  \State Draw a mixed batch; let $\mathcal V_t$ be its loss-bearing target
    positions with target-aligned domain IDs
  \State $q_{k,t}\gets \operatorname{CE}_{\theta_t}(y_k\mid x_{\leq k})$
    for every $k\in\mathcal V_t$
  \State $\mathcal L_t\gets
    |\mathcal V_t|^{-1}\sum_{k\in\mathcal V_t}w_{d_k,t}q_{k,t}$
  \State $\theta_{t+1}\gets
    \Call{OptimizerStep}{\mathcal O,\theta_t,\nabla_{\theta_t}\mathcal L_t}$
  \State Compute $S_{i,t}=\sum_{k\in\mathcal V_t}
    \mathbf{1}[d_k=i]q_{k,t}$ and
    $N_{i,t}=\sum_{k\in\mathcal V_t}\mathbf{1}[d_k=i]$ for all $i$
  \For{each domain $i\in\mathcal D$}
    \State $\hat{c}_{i,t}\gets
      \beta\hat{c}_{i,t-1}+(1-\beta)N_{i,t}$
    \If{$N_{i,t}>0$}
      \State $\bar{\ell}_{i,t}\gets S_{i,t}/N_{i,t}$
      \If{$\mathsf{seen}_i=\mathsf{false}$}
        \State $\hat{\ell}_{i,t}\gets\bar{\ell}_{i,t}$;
          $\mathsf{seen}_i\gets\mathsf{true}$
      \Else
        \State $\hat{\ell}_{i,t}\gets
          \beta\hat{\ell}_{i,t-1}+(1-\beta)\bar{\ell}_{i,t}$
      \EndIf
    \Else
      \State $\hat{\ell}_{i,t}\gets\hat{\ell}_{i,t-1}$
    \EndIf
  \EndFor
  \If{$t<T_{\mathrm w}$}
    \State $w_{i,t+1}\gets 1$ for all $i\in\mathcal D$
  \Else
    \State $\hat{\pi}_{i,t}\gets
      \hat{c}_{i,t}/\sum_{j\in\mathcal D}\hat{c}_{j,t}$
      for all $i\in\mathcal D$
    \State $r_{i,t}\gets(\hat{\ell}_{i,t}+\epsilon)^{-1/2}$ for each seen
      domain; $r_{i,t}\gets0$ otherwise
    \State $Z_t\gets\sum_{j\in\mathcal D}\hat{\pi}_{j,t}r_{j,t}$
    \State $w_{i,t+1}\gets r_{i,t}/Z_t$ for each seen domain;
      $w_{i,t+1}\gets1$ otherwise
  \EndIf
\EndFor
\end{algorithmic}
\end{algorithm}

\input{sections/90_appendix_theory/appendix_v3}

%% file: sections/90_appendix_theory/appendix_v3.tex
\subsection{Proof of Proposition~\ref{prop:v3-main}}
\label{app:proof-v3-main}

\paragraph{Model and SGD update.}
Let \(A_1,A_2\in\R^{d\times d}\) be nonzero symmetric positive semidefinite matrices with orthogonal supports, so \(A_1A_2=0\). In domain \(j\in\{1,2\}\), samples follow \(x_j\sim\mathcal N(0,A_j)\) and \(y_j=x_j^\top w^\star\). A shared \(w^\star\) is without loss of generality: ground-truth parameters that differ between domains can be projected onto their respective orthogonal supports and summed. The population MSE loss is
\[
 f_j(w)=\tfrac12\,\E_{x_j}\!\bigl[(x_j^\top w-y_j)^2\bigr]
       =\tfrac12\,(w-w^\star)^\top A_j(w-w^\star).
\]
Fix an initialization $w_0$ with $f_1(w_0),f_2(w_0)>0$, a batch size $B\geq2$, and a horizon $T\geq1$. At each step $t$, the batch contains exactly $Bp_j\in\mathbb N^+$ fresh samples from domain $j$, where $p_j\in(0,1)$ and $p_1+p_2=1$. Samples are independent across domains, examples, and iterations. With per-domain learning rates $\eta_1,\eta_2>0$, SGD updates
\begin{equation*}
 w_{t+1}=w_t-\frac1B\sum_{j=1}^2\eta_j
 \sum_{i=1}^{Bp_j}x_{j,i,t}
 \bigl(x_{j,i,t}^\top w_t-y_{j,i,t}\bigr).
\end{equation*}
We write $L_j(p_j,\eta_j)=\E[f_j(w_T)]$ for the expected terminal loss of domain $j$.

The ranges of $A_1$ and $A_2$ are orthogonal since $A_1A_2=0$.
A domain-$j$ sample $x_j\sim\mathcal N(0,A_j)$ lies in the range of
$A_j$ almost surely, so its stochastic gradient $x_j(x_j^\top w_t-y_j)$ does too.
A domain-$j$ update therefore changes only the components of the error
$w_t-w^\star$ that lie in the range of $A_j$.
Consequently, the two domains never interact, and
the terminal loss $L_j$ depends only on that domain's own mixture proportion $p_j$ and learning rate $\eta_j$. We therefore prove the single-domain
Lemmas~\ref{lem:v3-recursion} and~\ref{lem:v3-monotonicity} with the domain index
dropped from the notation, and apply them to $j=1,2$ in the proof of the
proposition.

\paragraph{Single-domain notation.}
Fix one domain and drop its index. Let $r=\rank(A_j)$ be the dimension of its
positive support, and diagonalize the covariance restricted to that support as
\begin{equation}
 D=\diag(\lambda_1,\ldots,\lambda_r),
 \qquad \lambda=(\lambda_1,\ldots,\lambda_r)^\top>0,
 \qquad z_{t,k}=\E[e_{t,k}^2],
 \label{eq:v3-single-domain-notation}
\end{equation}
where $\lambda_1,\ldots,\lambda_r>0$ are the positive eigenvalues,
$e_t=w_t-w^\star$ is the error, $e_{t,k}$ (for $k=1,\ldots,r$) is its coordinate
along the $k$-th eigenvector, and $z_t=(z_{t,1},\ldots,z_{t,r})^\top$ collects the
per-coordinate second moments. Since $\E[f(w_t)]=\tfrac12\,\E[e_t^\top De_t]
=\tfrac12\lambda^\top z_t$, positive initial domain loss is exactly
$\lambda^\top z_0>0$.

\begin{lemma}
\label{lem:v3-recursion}
Suppose the domain receives a fraction $p\in(0,1)$ of each batch (that is, $Bp$ of
the $B$ samples) and learning rate $\eta\geq0$. Then the second-moment vector
$z_t$ and the expected terminal loss satisfy
\begin{align}
 z_{t+1}&=K(p,\eta)\,z_t,\nonumber\\
 K(p,\eta)&=(I-p\eta D)^2
 +\frac{p\eta^2}{B}\bigl(D^2+\lambda\lambda^\top\bigr),
 \label{eq:v3-moment-matrix}\\
 L(p,\eta)&=\tfrac12\,\lambda^\top[K(p,\eta)]^{\,T}z_0.
 \label{eq:v3-exact-loss}
\end{align}
\end{lemma}

\begin{proof}
Write $n=Bp$ for the number of domain samples in a batch. 
SGD updates
\begin{equation}
 e_{t+1}
 =e_t-\frac{\eta}{B}\sum_{i=1}^{n}x_i\bigl(x_i^\top w_t-y_i\bigr)
 =\Bigl(I-\frac{\eta}{B}\sum_{i=1}^{n}x_ix_i^\top\Bigr)e_t,
 \qquad x_i\sim\mathcal N(0,D)\ \text{i.i.d.}
 \label{eq:v3-error-map}
\end{equation}
Condition on $e_t=e$ and set $g_i=x_i(x_i^\top e)$.
For a
mean-zero Gaussian with diagonal covariance $D$, Isserlis' theorem gives
$\E[x_k^2x_\ell x_m]=\lambda_k\lambda_\ell\,\delta_{\ell m}
+2\lambda_k^2\,\delta_{k\ell}\delta_{km}$, whence, coordinatewise,
\begin{equation}
 \E[g_{i,k}]=\lambda_ke_k,
 \qquad
 \Var(g_{i,k})=\E[g_{i,k}^2]-(\E[g_{i,k}])^2
 =\lambda_k^2e_k^2+\lambda_k\sum_{\ell=1}^r\lambda_\ell e_\ell^2.
 \label{eq:v3-gaussian-moments}
\end{equation}
Coordinate $k$ of \eqref{eq:v3-error-map} is $e_{t+1,k}=e_k-\frac{\eta}{B}S_k$ with
$S_k=\sum_{i=1}^n g_{i,k}$. Its first two moments are $\E[S_k]=n\lambda_ke_k$ and
$\E[S_k^2]=\Var(S_k)+(\E[S_k])^2=n\Var(g_{1,k})+n^2\lambda_k^2e_k^2$, so, expanding
the square and using $n/B=p$,
\begin{align}
 \E[e_{t+1,k}^2\mid e_t=e]
 &=e_k^2-\frac{2\eta}{B}\,e_k\,\E[S_k]+\frac{\eta^2}{B^2}\,\E[S_k^2]\nonumber\\
 &=e_k^2-2p\eta\lambda_ke_k^2+p^2\eta^2\lambda_k^2e_k^2
   +\frac{p\eta^2}{B}\Var(g_{1,k})\nonumber\\
 &=(1-p\eta\lambda_k)^2e_k^2
   +\frac{p\eta^2}{B}\Bigl(\lambda_k^2e_k^2
   +\lambda_k\sum_{\ell}\lambda_\ell e_\ell^2\Bigr).
 \label{eq:v3-coordinate-recursion}
\end{align}
Taking the expectation over $e_t$ yields 
\begin{equation}
    z_{t+1}=K(p, \eta)z_t.
\end{equation}
Iterating gives
$z_T=[K(p,\eta)]^Tz_0$, and since $f(w_T)=\tfrac12 e_T^\top De_T
=\tfrac12\sum_k\lambda_ke_{T,k}^2$,
\[
 L(p,\eta)=\E[f(w_T)]=\tfrac12\lambda^\top z_T
 =\tfrac12\lambda^\top[K(p,\eta)]^Tz_0,
\]
which is \eqref{eq:v3-exact-loss}.
\end{proof}

\begin{lemma}
\label{lem:v3-monotonicity}
For fixed $B,T$, let $V(p)=\min_{\eta\geq0}L(p,\eta)$. The minimum is attained,
every minimizing rate is positive, and $V(p)$ is strictly decreasing in
$p\in(0,1)$.
\end{lemma}

\begin{proof}
All matrix inequalities in this proof are entrywise. We read $L(p,\eta)$ off the
closed form~\eqref{eq:v3-exact-loss} for every $p\in(0,1)$; at a feasible mixture
($Bp\in\mathbb N$) it is the SGD loss of Lemma~\ref{lem:v3-recursion}. Set the
effective mean step $a=p\eta$ and define
\begin{equation}
 \widehat K_p(a)=(I-aD)^2+\frac{a^2}{Bp}\bigl(D^2+\lambda\lambda^\top\bigr),
 \qquad
 F_p(a)=\tfrac12\,\lambda^\top[\widehat K_p(a)]^{\,T}z_0,
 \label{eq:v3-effective-step}
\end{equation}
so that $F_p(p\eta)=L(p,\eta)$; since $p>0$, minimizing $F_p$ over $a\geq0$ is
equivalent to minimizing $L(p,\cdot)$ over $\eta\geq0$. The map $F_p$ is a
polynomial, hence continuous, and its derivative at zero is
\begin{equation}
 F_p'(0)=-T\sum_{k=1}^r\lambda_k^2z_{0,k}<0.
 \label{eq:v3-negative-derivative}
\end{equation}
It is also coercive: $\widehat K_p(a)\geq(a^2/(Bp))\,\lambda\lambda^\top$, and
\begin{equation}
 F_p(a)\geq \tfrac12\lambda^\top z_0 \Bigl(\frac{a^2\lambda^\top\lambda}{Bp}\Bigr)^{T}\longrightarrow\infty
 \qquad\text{as }a\to\infty.
 \label{eq:v3-coercivity}
\end{equation}
Continuity and coercivity give a nonempty compact set of minimizers, while
\eqref{eq:v3-negative-derivative} excludes $a=0$.
So every minimizing rate is positive.

Now take $0<p<\tilde p<1$ and fix any $a>0$. With $U=\widehat K_{p}(a)$ and
$W=\widehat K_{\tilde p}(a)$,
\begin{equation}
 U-W=\frac{a^2}{B}\Bigl(\frac1{p}-\frac1{\tilde p}\Bigr)
 \bigl(D^2+\lambda\lambda^\top\bigr)>0,
 \label{eq:v3-strict-matrix-order}
\end{equation}
and both $U,W$ have strictly positive entries. The noncommutative telescoping
identity
\begin{equation}
 U^T-W^T=\sum_{k=0}^{T-1}U^{\,T-1-k}(U-W)\,W^{k}>0
 \label{eq:v3-telescope}
\end{equation}
then gives $F_{\tilde p}(a)<F_{p}(a)$. Taking any minimizer $a_\star>0$ of $F_p$,
\begin{equation}
 V(\tilde p)\leq F_{\tilde p}(a_\star)<F_{p}(a_\star)=V(p),
 \label{eq:v3-strict-value-decrease}
\end{equation}
which proves strict decrease.
\end{proof}

\begin{proof}[Proof of Proposition~\ref{prop:v3-main}]
By the reduction above, $L_j(p_j,\eta_j)$ is the single-domain loss of
Lemma~\ref{lem:v3-recursion} for the domain-$j$ spectrum, and
$V_j(p)=\min_{\eta\geq0}L_j(p,\eta)$ its optimized value. By
Lemma~\ref{lem:v3-monotonicity} each $V_j$ is attained at a positive rate, so the
reweighting rates $\eta_j^\star$, which minimize over $\eta_j>0$, achieve it:
\begin{equation}
 L_1(p_1,\eta_1^\star)=V_1(p_1),
 \qquad
 L_2(p_2,\eta_2^\star)=V_2(p_2).
 \label{eq:v3-reweight-point}
\end{equation}
Consider any remix with mixture $(p_1',p_2')\neq(p_1,p_2)$ and shared rate
$\eta'$; since $p_1'+p_2'=1=p_1+p_2$, this is equivalent to $p_1'\neq p_1$. Because
$V_j(p_j')$ is the minimum of $L_j(p_j',\cdot)$,
\begin{equation}
 L_1(p_1',\eta')\geq V_1(p_1'),
 \qquad
 L_2(p_2',\eta')\geq V_2(p_2').
 \label{eq:v3-remix-lower-bound}
\end{equation}
If $p_1'<p_1$, strict monotonicity of $V_1$ (Lemma~\ref{lem:v3-monotonicity})
with~\eqref{eq:v3-remix-lower-bound} gives
\begin{equation}
 L_1(p_1',\eta')\geq V_1(p_1')>V_1(p_1)=L_1(p_1,\eta_1^\star).
 \label{eq:v3-left-obstruction}
\end{equation}
If instead $p_1'>p_1$, the same argument applied
to domain $2$ gives
\begin{equation}
 L_2(p_2',\eta')\geq V_2(p_2')>V_2(p_2)=L_2(p_2,\eta_2^\star).
 \label{eq:v3-right-obstruction}
\end{equation}
In either case one coordinate of the remix strictly exceeds the corresponding
coordinate of the reweighting point. This is exactly the
separation~\eqref{eq:v3-remix-separation}, and it shows that no remix
Pareto-dominates $\bigl(L_1(p_1,\eta_1^\star),L_2(p_2,\eta_2^\star)\bigr)$.
\end{proof}

\subsection{Numerical instantiation of Figure~\ref{fig:quadratic-pareto-separation}}
\label{app:v3-numerics}

Figure~\ref{fig:quadratic-pareto-separation} evaluates the two-domain quadratic
model of Proposition~\ref{prop:v3-main} on a fixed, non-degenerate instance.

\paragraph{Model instance.} We use batch size $B=128$, horizon $T=20$, and two
orthogonal coordinate blocks of dimension $d_1=d_2=512$. The covariance $A_j$ is
diagonal on block $j$, with eigenvalues drawn once, independently, as
\begin{equation}
 \lambda_{j,k}=a_j\,u_{j,k},
 \qquad u_{j,k}\overset{\text{i.i.d.}}{\sim}\mathrm{Uniform}(0.5,1.5),
 \qquad k=1,\dots,d_j,
 \label{eq:v3-random-spectrum}
\end{equation}
with block scales $a_1=2.5$ and $a_2=1$ and a fixed random seed. Each domain
therefore has mean curvature $a_j$ and a genuine spread (within-block condition
number $3$), rather than the degenerate flat spectrum $\lambda_{j,k}\equiv a_j$;
the scale gap between the blocks is what makes the two domains' optimal rates
differ. We start from an isotropic error, $z_0=\mathbf 1$ (unit expected squared
error in every eigen-coordinate), so the initial loss
$f_j(w_0)=\tfrac12\lambda_j^\top\mathbf 1>0$ and the hypothesis of
Proposition~\ref{prop:v3-main} holds.

\paragraph{Computing the expectation.} All losses are computed exactly, with no
Monte Carlo. For a domain with spectrum $D=\diag(\lambda)$, batch fraction $p$,
and rate $\eta$, we iterate the second-moment
recursion~\eqref{eq:v3-moment-matrix} of Lemma~\ref{lem:v3-recursion} in its
coordinate form,
\begin{equation}
 z_{t+1,k}
 =\Bigl[(1-p\eta\lambda_k)^2+\tfrac{p\eta^2}{B}\lambda_k^2\Bigr]z_{t,k}
 +\frac{p\eta^2}{B}\,\lambda_k\sum_{\ell}\lambda_\ell z_{t,\ell},
 \label{eq:v3-coordinate-iteration}
\end{equation}
for $t=0,\dots,T-1$ starting from $z_0=\mathbf 1$, and read off
$L(p,\eta)=\tfrac12\lambda^\top z_T$ via~\eqref{eq:v3-exact-loss}. The figure
plots the normalized loss $L(p,\eta)/L(p,0)=\lambda^\top z_T/\lambda^\top z_0$,
which equals $\E[f_j(w_T)]$ up to the fixed positive factor $f_j(w_0)$ and so
preserves every Pareto comparison. Because the transition is
diagonal-plus-rank-one, each of the $T$ steps costs $O(d_j)$, and the whole rate
grid below is evaluated by broadcasting~\eqref{eq:v3-coordinate-iteration} over
all rates at once.

\paragraph{Sweeping the learning rate.} Once the spectrum is non-flat, the
minimizing rates have no closed form---Lemma~\ref{lem:v3-monotonicity} guarantees
a positive minimizer but not its value---so we optimize by sweeping. We evaluate
$L(p,\eta)$ on a uniform grid of $2000$ rates in $[0,\eta_{\max}]$, with
$\eta_{\max}=1.5\big/\bigl(\lambda_{\min}\,(p_{\min}+(d+1)/B)\bigr)$; this
inflates the flat-spectrum optimum evaluated at the smallest present eigenvalue
$\lambda_{\min}$ and the smallest batch fraction $p_{\min}=8/B$, and we check
that no reported optimum falls on either endpoint. Mixtures are $p_1=n_1/B$ with
$n_1\in\{8,\dots,120\}$, so that $Bp_j\in\mathbb N^+$. For each mixture
$(p_1,p_2)$:
\begin{itemize}
 \item the \emph{reweighting} point is
   $\bigl(\min_\eta L_1(p_1,\eta),\ \min_\eta L_2(p_2,\eta)\bigr)$, minimizing
   each coordinate independently over the grid (the sweep analogue of the
   $V_j(p_j)$ in the proof);
 \item the \emph{remixing} attainable set collects
   $\bigl(L_1(p_1,\eta),L_2(p_2,\eta)\bigr)$ over all grid rates $\eta$ and all
   mixtures, and its lower-left envelope is the shared-rate Pareto frontier.
\end{itemize}
The left panel fixes $p_1=\tfrac12$: the reweighting point uses the two
per-domain optimal rates, while the shared baseline is the single rate
minimizing the mixture-weighted loss $p_1L_1+p_2L_2$. The right panel overlays
the reweighting frontier (the reweighting point traced over all mixtures) on the
shared-rate frontier.

%% file: sections/91_appendix_data/section.tex
\section{Data Sources and Processing}
\label{app:data}

We use three data domains: general knowledge (K), mathematics (M), and code (C). All corpora are tokenized with the GPT-2 BPE tokenizer, and every stored token carries an aligned loss mask and domain ID (0 for K, 1 for M, and 2 for C). This appendix describes the source pools, filtering rules, QA conversion, and the distinction between sampled-token exposure and loss-bearing tokens.

\subsection{Sources and Filtering}
\label{app:data-sources}

\paragraph{Knowledge.} We use full-text web documents~\citep{penedo2024fineweb} as the knowledge-domain corpus. The prepared pool contains 289.27B tokens.

\paragraph{Mathematics.} We collect mathematical problems and worked solutions from multiple sources~\citep{li2023camel,tigerlab2024mathplus,yue2023mammoth,yu2023metamath,liu2024augmenting,li2024numinamath,toshniwal2024openmath2,openr1team2025openr1math220k,mitra2024orcamath,fujii2025rewritingpretrainingdataboosts,yue2024mammoth2}. We first remove examples with 13-token n-gram overlap against GSM8K~\citep{cobbe2021gsm8k}, MATH~\citep{hendrycks2021math}, AIME 2024~\citep{AIME2024}, AIME 2025~\citep{AIME2025}, HMMT 2025~\citep{HMMT2025}, AMC 2023~\citep{AMC2023}, or the English split of OlympiadBench~\citep{he2024olympiadbench}. We then deduplicate normalized $(\text{question},\text{answer})$ rows in a fixed source order, retaining the first occurrence. This conservative rule preserves distinct solutions to the same question. After filtering and deduplication, the prepared pool contains 17.20B tokens.

\paragraph{Code.} We collect code-oriented prompts, responses, and source files from multiple sources~\citep{nvidia2025nemotronnano2,huang2024opencoder,fujii2025rewritingpretrainingdataboosts}. We remove examples with 10-token n-gram overlap against HumanEval~\citep{chen2021humaneval}, MBPP~\citep{austin2021mbpp}, GSM8K~\citep{cobbe2021gsm8k}, or MATH~\citep{hendrycks2021math}. We then perform exact cross- and intra-source deduplication on code answers of at least 200 characters, followed by exact $(\text{question},\text{answer})$ row deduplication at every length. The final training pool contains 245.16B tokens.

\begin{table}[ht!]
  \caption{\small\textbf{Prepared source pools for the 4$\times$/8$\times$ experiments.} Counts are pre-mask GPT-2 BPE tokens; code excludes source-level validation files.}
  \label{tab:dataset-summary}
  \begin{center}
  \small
  \begin{tabular}{cllr}
    \toprule
    ID & Domain & Representation & Tokens \\
    \midrule
    0 & Knowledge & Full text & 289.27B \\
    1 & Mathematics & QA pairs & 17.20B \\
    2 & Code & QA pairs & 245.16B \\
    \bottomrule
  \end{tabular}
  \end{center}
\end{table}

\subsection{Training Records and Loss Masking}
\label{app:qa-contract}

\paragraph{Record format.} After source-specific processing, every mathematics or code example is serialized as $[q,a]$, where $q$ is the prompt and $a$ is the target response. Any record whose complete serialization exceeds 4,096 GPT-2 BPE tokens is discarded.

\paragraph{Mathematics.} The source problem forms $q$. After a blank line, we append the following fixed instruction:

\begin{center}
\fbox{\parbox{0.88\linewidth}{\centering\small\ttfamily Please put the answer in \textbackslash boxed\{\}.}}
\end{center}

The target response $a$ retains the worked solution supplied by the source when one is available and always ends with a normalized $\texttt{\textbackslash boxed\{...\}}$ line containing the final answer. Rows from which no reliable final answer can be extracted are omitted.

\paragraph{Code.} We parse each code source according to its native format and add no shared instruction. For the instruction corpus with explicit reasoning tags, $q$ is the text before \texttt{<think>} and $a$ is the complete response after \texttt{</think>}; the intervening reasoning trace is removed. For corpora organized around fenced code blocks, we split at the first fenced code block: the preceding text becomes $q$, while the fence and everything after it become $a$. For the corpus of bare source files, each file is wrapped in a Python code fence and stored with an empty $q$, reducing the record to $[a]$.

\paragraph{Loss masks.} For mathematics and code, the stored mask is zero on \textsc{eot} and prompt tokens and one on response tokens. For full-text knowledge, it is zero on document-separating \textsc{eot} tokens and one on document text. During training, a second mask suppresses the first 100 prediction targets after every \textsc{eot} encountered within a loader window. The distance from a preceding \textsc{eot} is computed only within that window and is not carried across window boundaries. The effective loss mask is the intersection of the stored and training-time masks.

Masks and domain IDs are attached to prediction targets. Thus, for the causal pair $(x_t,x_{t+1})$, the loss reads the mask and domain ID associated with $x_{t+1}$. A masked target contributes neither to the language-model loss nor to the per-domain statistics used by \Method{}. Records are concatenated into a padding-free stream, and the token, mask, and domain-ID arrays remain positionally aligned.

\subsection{Mixture Ratios and Realized Loss-Token Shares}
\label{app:data-mixtures}

Every mixture ratio in the paper is defined over sampled tokens before loss masking. For each declared ratio and token budget, we construct one frozen training stream and reuse it for every controlled comparison at that setting. Within every controlled comparison, runs use exactly the same frozen training stream, including sample order and loss masks; only the experimental factor under study changes. This applies to the optimizer comparisons and every direct--\Method{} pair. Experiments with different ratios or token budgets use their corresponding frozen streams.

\begin{table*}[ht!]
  \caption{\small\textbf{Sampled-token exposure and realized loss-token shares.} Shares are means over per-step valid-target counts after the on-disk and 100-target burn-in masks.}
  \label{tab:mixture-audit}
  \begin{center}
  \small
  \begin{tabular}{llrrrrrr}
    \toprule
    & & \multicolumn{3}{c}{Sampled tokens (B)} & \multicolumn{3}{c}{Loss-token share (\%)} \\
    \cmidrule(lr){3-5}\cmidrule(lr){6-8}
    Budget & Sampled ratio & K & M & C & K & M & C \\
    \midrule
    1$\times$ & $7{:}1$ K:M & 26.25 & 3.75 & -- & 90.07 & 9.93 & -- \\
    1$\times$ & $3{:}1$ K:M & 22.50 & 7.50 & -- & 79.53 & 20.47 & -- \\
    1$\times$ & $1{:}1$ K:M & 15.00 & 15.00 & -- & 56.42 & 43.58 & -- \\
    \addlinespace
    1$\times$ & $7{:}0.5{:}0.5$ K:M:C & 26.25 & 1.875 & 1.875 & 90.28 & 4.97 & 4.75 \\
    1$\times$ & $3{:}0.5{:}0.5$ K:M:C & 22.50 & 3.75 & 3.75 & 79.69 & 10.25 & 10.06 \\
    1$\times$ & $1{:}0.5{:}0.5$ K:M:C & 15.00 & 7.50 & 7.50 & 56.75 & 21.91 & 21.34 \\
    \midrule
    4$\times$ & $7{:}1$ K:M & 105.00 & 15.00 & -- & 90.07 & 9.93 & -- \\
    4$\times$ & $7{:}0.5{:}0.5$ K:M:C & 105.00 & 7.50 & 7.50 & 90.12 & 4.97 & 4.91 \\
    8$\times$ & $7{:}0.5{:}0.5$ K:M:C & 210.00 & 15.00 & 15.00 & 90.11 & 4.97 & 4.92 \\
    \bottomrule
  \end{tabular}
  \end{center}
\end{table*}

Table~\ref{tab:mixture-audit} reports both the nominal sampled-token exposure and the mean domain share among valid loss targets after applying the on-disk and 100-target burn-in masks. Masking changes the effective composition: for example, a nominal $1{:}1$ K:M mixture contains only 43.58\% mathematics tokens among the targets that contribute to the loss. Sampled-token ratios should therefore be interpreted as data exposure rather than as relative contributions to the unweighted objective. Token allocations use the rounded 1$\times$/4$\times$/8$\times$ budgets of 30B/120B/240B.

%% file: sections/93_appendix_details/section.tex
\section{Complete Experimental Details and Results}
\label{app:details}

\subsection{Detailed Hyperparameters}
\label{app:hyperparameters}

\paragraph{Model and training.} The main experiments use a 1.635B-parameter decoder-only Transformer with 48 layers, width 1600, 25 attention heads, and a 50,304-token vocabulary. All linear maps are bias-free, normalization is parameter-free, and rotary position encoding is used. The 1.5B runs use a 4096-token context, a global batch of 128 sequences, bfloat16 autocasting and linear learning-rate warmup followed by cosine decay over the full training budget. In multi-stage runs, one global learning-rate schedule spans all stages: the warmup and cosine decay do not restart at a stage boundary, and the optimizer state carries over. If the first stage ends before warmup does, warmup continues into the next stage. Each optimizer step consumes 524,288 sampled tokens, yielding 29.9998B, 119.9990B, and 239.9981B tokens after 57,220, 228,880, and 457,760 steps, respectively. Table~\ref{tab:model-training-overview} summarizes the experiment blocks.

\begin{table}[ht!]
  \caption{\small Model, token budget, data composition, and optimizer for each experiment block. Ratios refer to sampled tokens before loss masking.}
  \label{tab:model-training-overview}
  \begin{center}
  \resizebox{\linewidth}{!}{%
  \begin{tabular}{llrrll}
    \toprule
    Block & Model & Tokens & Steps & Data ratio & Optimizer(s) \\
    \midrule
    1$\times$ controlled experiments & 1.635B & 30B & 57,220 & three K:M or K:M:C ratios & AdamW/Muon/\agd{} \\
    4$\times$ K:M & 1.635B & 120B & 228,880 & 7:1 & \agd{} \\
    4$\times$ K:M:C & 1.635B & 120B & 228,880 & 7:0.5:0.5 & \agd{} \\
    8$\times$ K:M:C & 1.635B & 240B & 457,760 & 7:0.5:0.5 & \agd{} \\
    Model scaling & 124M/350M/774M & 2.52B/7.08B/15.73B & 4,800/13,500/30,000 & 7:1 & \agd{} \\
    \bottomrule
  \end{tabular}}
  \end{center}
\end{table}

\paragraph{Optimizers.} AdamW uses $(\beta_1,\beta_2)=(0.9,0.98)$ and $\epsilon=10^{-8}$. Muon uses momentum 0.9 with Nesterov acceleration and five Newton--Schulz iterations for matrix-valued hidden parameters, together with auxiliary AdamW for the embedding and output matrices. \agd{} uses four Newton--Schulz iterations, $(\beta_1,\beta_2)=(0.95,0.95)$, $\epsilon=10^{-5}$, and the row-wise $(2,\infty)$ direction for both vocabulary-facing and transformer-block parameters; Appendix~\ref{app:agd} gives the complete update. Learning rates are matched within every direct--adaptive pair, and each reported cell is a single training run.

\begin{table}[ht!]
  \caption{\small Optimizer and learning-rate schedule settings. Muon uses separate learning rates for Muon matrices and auxiliary AdamW parameters; the model-scaling warmups correspond to 124M, 350M, and 774M.}
  \label{tab:hyperparameters}
  \begin{center}
  \resizebox{\linewidth}{!}{%
  \begin{tabular}{lllll}
    \toprule
    Block & Optimizer & Peak LR & Weight decay & Warmup \\
    \midrule
    30B AdamW experiments & AdamW & 0.001 & 0.10 & 20,000 \\
    30B Muon experiments & Muon + AdamW & 0.005 / 0.001 & 0.10 & 20,000 \\
    30B \agd{} experiments & \agd{} & 0.010 & $10^{-5}$ & 20,000 \\
    120B/240B main experiments & \agd{} & 0.010 & $10^{-5}$ & 20,000 \\
    Model scaling & \agd{} & 0.010 & $10^{-5}$ & 1,678 / 4,719 / 10,486 \\
    \bottomrule
  \end{tabular}}
  \end{center}
\end{table}

\paragraph{Controller.} All \Method{} runs follow Algorithm~\ref{alg:online-invsqrt} with $\beta=0.99$, $\epsilon=10^{-8}$, and the learning-rate warmup lengths in Table~\ref{tab:hyperparameters}. At each optimizer step, the controller aggregates each domain's unweighted cross-entropy sum and valid-target count over the global batch, updates the count EMA for every domain and the loss EMA for each observed domain, and keeps all loss coefficients equal to one during warmup. After warmup, it computes $r_{i,t}=(\hat{\ell}_{i,t}+\epsilon)^{-1/2}$ and applies $w_{i,t+1}=r_{i,t}/\sum_j\hat{\pi}_{j,t}r_{j,t}$, where $\hat{\pi}_{j,t}$ is the EMA count share. Fixed-weight ablations likewise use unit weights during warmup and activate their listed coefficients afterward.

\paragraph{Reproducibility.} Within every controlled comparison, the paired runs use the same prepared data streams described in Appendix~\ref{app:data}. Per-domain loss sums and valid-token counts are aggregated across data-parallel ranks before the controller is updated, so every rank applies the same weights on the next optimizer step. All tables report the unsmoothed final checkpoint; any smoothing in a training-curve figure is used only for display.

\subsection{Matched-Exposure Results}
\label{app:phenomenon-results}

Table~\ref{tab:phenomenon} compares four 30B-token recipes across AdamW, Muon, and \agd{} with 26.25B knowledge and 3.75B mathematics tokens. Figure~\ref{fig:phenomenon-curves} shows the Muon checkpoint trajectories.

\begin{table}[ht!]
  \caption{\small Matched-exposure comparison at 30B tokens across optimizers. First-domain drop is the final score minus the best pre-switch score; bold marks the best reported final result within each optimizer and benchmark.}
  \label{tab:phenomenon}
  \begin{center}
  \resizebox{\linewidth}{!}{%
  \begin{tabular}{llllccc}
    \toprule
    Optimizer & Recipe & Token schedule & $w_{\mathrm{K}}:w_{\mathrm{M}}$ & MMLU & GSM8K & First-domain drop \\
    \midrule
    \multirow{4}{*}{AdamW}
      & Knowledge $\rightarrow$ Math & 26.25B K $\rightarrow$ 3.75B M & 1:1 & 30.57 & 18.35 & MMLU: $-1.06$ \\
      & Math $\rightarrow$ Knowledge & 3.75B M $\rightarrow$ 26.25B K & 1:1 & \textbf{32.28} & 0.83 & GSM8K: $-7.96$ \\
      & Direct mixture & stationary 7:1 & 1:1 & 31.30 & 21.91 & -- \\
      & Fixed reweight & stationary 7:1 & 1:2 & 32.03 & \textbf{31.24} & -- \\
    \midrule
    \multirow{4}{*}{Muon}
      & Knowledge $\rightarrow$ Math & 26.25B K $\rightarrow$ 3.75B M & 1:1 & 29.45 & 36.62 & MMLU: $-1.63$ \\
      & Math $\rightarrow$ Knowledge & 3.75B M $\rightarrow$ 26.25B K & 1:1 & 31.78 & 0.76 & GSM8K: $-17.74$ \\
      & Direct mixture & stationary 7:1 & 1:1 & 31.40 & 27.67 & -- \\
      & Fixed reweight & stationary 7:1 & 1:2 & \textbf{32.27} & \textbf{38.59} & -- \\
    \midrule
    \multirow{4}{*}{\agd{}}
      & Knowledge $\rightarrow$ Math & 26.25B K $\rightarrow$ 3.75B M & 1:1 & 30.97 & 11.45 & MMLU: $-1.70$ \\
      & Math $\rightarrow$ Knowledge & 3.75B M $\rightarrow$ 26.25B K & 1:1 & 31.95 & 0.61 & GSM8K: $-26.23$ \\
      & Direct mixture & stationary 7:1 & 1:1 & 31.82 & 28.58 & -- \\
      & Fixed reweight & stationary 7:1 & 1:2 & \textbf{32.65} & \textbf{34.19} & -- \\
    \bottomrule
  \end{tabular}}
  \end{center}
\end{table}

Across all three optimizers, the two-stage orders retain different capabilities. Fixed $1{:}2$ reweighting improves both MMLU and GSM8K over direct mixing under every optimizer. The Muon reweighted endpoint exceeds the best scores along either stage trajectory; under AdamW, the Math $\rightarrow$ Knowledge run retains a higher MMLU score.

\subsection{Detailed Data-Mixture Results}
\label{app:ratio-results}

\begin{figure*}[ht!]
  \centering
  \includegraphics[width=\textwidth]{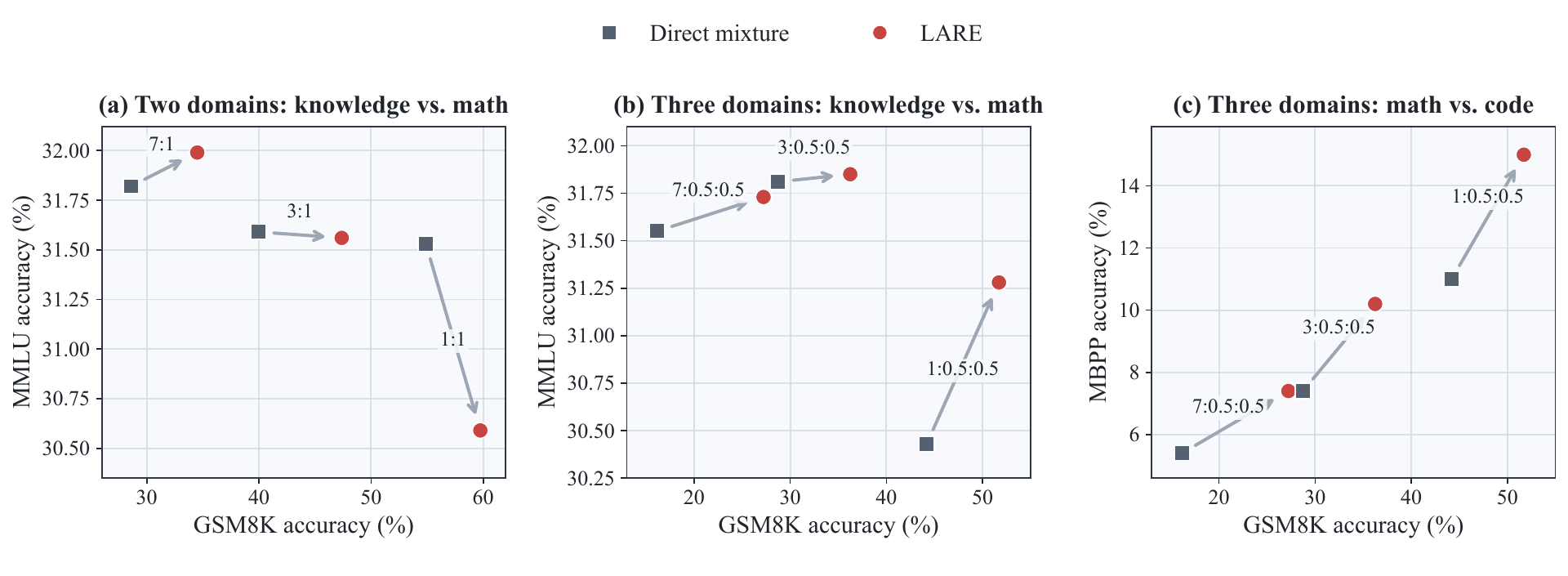}
  \caption{\textbf{\Method{} remains effective across mixture ratios and improves all three capabilities in K:M:C training.} Arrows connect matched direct-mixture and \Method{} runs with the same $1\times$ Chinchilla budget and \agd{} configuration. Panel (a) varies K:M over $7{:}1$, $3{:}1$, and $1{:}1$: GSM8K improves at every ratio, while the largest MMLU cost occurs at the reasoning-heavy $1{:}1$ boundary. Panels (b) and (c) add code with M:C $=1{:}1$; \Method{} improves MMLU, GSM8K, and MBPP at every tested ratio.}
  \label{fig:ratio-pareto}
\end{figure*}

Figure~\ref{fig:ratio-pareto} visualizes the \agd{} data-mixture comparisons, and Table~\ref{tab:ratio-ablation} reports their final checkpoints.

\begin{table*}[ht!]
  \caption{\small Direct and adaptive objectives across 30B-token data mixtures with \agd{}. Each pair differs only in online loss reweighting; bold marks the better result within the pair, and ``--'' indicates that the corresponding domain is absent.}
  \label{tab:ratio-ablation}
  \begin{center}
  \begin{tabular}{lllrrr}
    \toprule
    Domains & Data ratio & Objective & MMLU & GSM8K & MBPP \\
    \midrule
    \multirow{6}{*}{K:M} & \multirow{2}{*}{7:1} & Direct mixture & 31.82 & 28.58 & -- \\
      & & \Method{} & \textbf{31.99} & \textbf{34.50} & -- \\
    \cmidrule(lr){2-6}
      & \multirow{2}{*}{3:1} & Direct mixture & \textbf{31.59} & 39.95 & -- \\
      & & \Method{} & 31.56 & \textbf{47.38} & -- \\
    \cmidrule(lr){2-6}
      & \multirow{2}{*}{1:1} & Direct mixture & \textbf{31.53} & 54.89 & -- \\
      & & \Method{} & 30.59 & \textbf{59.74} & -- \\
    \midrule
    \multirow{6}{*}{K:M:C} & \multirow{2}{*}{7:0.5:0.5} & Direct mixture & 31.55 & 16.15 & 5.40 \\
      & & \Method{} & \textbf{31.73} & \textbf{27.22} & \textbf{7.40} \\
    \cmidrule(lr){2-6}
      & \multirow{2}{*}{3:0.5:0.5} & Direct mixture & 31.81 & 28.73 & 7.40 \\
      & & \Method{} & \textbf{31.85} & \textbf{36.24} & \textbf{10.20} \\
    \cmidrule(lr){2-6}
      & \multirow{2}{*}{1:0.5:0.5} & Direct mixture & 30.43 & 44.20 & 11.00 \\
      & & \Method{} & \textbf{31.28} & \textbf{51.71} & \textbf{15.00} \\
    \bottomrule
  \end{tabular}
  \end{center}
\end{table*}

At $1\times$ Chinchilla, \Method{} improves GSM8K by 4.85--7.43 points across K:M ratios $7{:}1$, $3{:}1$, and $1{:}1$; the corresponding MMLU changes are $+0.17$, $-0.03$, and $-0.94$ points (Figure~\ref{fig:ratio-pareto}(a)). At the reasoning-heavy $1{:}1$ boundary, mathematics contributes approximately 43\% of realized loss tokens but 68\% of the weighted objective at the final checkpoint, consistent with the larger MMLU cost. With code added at K:M:C ratios $7{:}0.5{:}0.5$, $3{:}0.5{:}0.5$, and $1{:}0.5{:}0.5$, \Method{} improves MMLU by 0.04--0.85 points, GSM8K by 7.51--11.07 points, and MBPP by 2.00--4.00 points at every ratio (Figure~\ref{fig:ratio-pareto}(b)--(c)). Additional optimizer replications appear in Appendix~\ref{app:optimizer-details}.

\subsection{Additional Optimizer Results}
\label{app:optimizer-details}

The ratio study referenced in Section~\ref{sec:ratio-ablation} and detailed in Appendix~\ref{app:ratio-results} uses \agd{}. To test whether its behavior is optimizer-specific, we repeat the same 30B-token K:M sweep with AdamW and Muon. Within each row of Table~\ref{tab:optimizer-ratio-replications}, the direct and \Method{} runs use the same model, prepared data stream, sampled-data ratio, optimizer settings, learning rate, and training budget; only online loss reweighting differs. The \agd{} endpoints remain in Table~\ref{tab:ratio-ablation} and are not duplicated here.

\begin{table}[ht!]
  \caption{\small Additional 30B-token K:M ratio sweeps with AdamW and Muon. Score pairs are MMLU/GSM8K accuracy at the final checkpoint; $\Delta$ is \Method{} minus direct mixture in percentage points.}
  \label{tab:optimizer-ratio-replications}
  \begin{center}
  \resizebox{0.8\linewidth}{!}{%
  \begin{tabular}{llccc}
    \toprule
    Optimizer & K:M & Direct mixture & \Method{} & $\Delta$ \\
    \midrule
    \multirow{3}{*}{AdamW} & 7:1 & 31.30 / 21.91 & 31.40 / 33.66 & $+0.10$ / $+11.75$ \\
      & 3:1 & 31.98 / 37.07 & 31.95 / 46.63 & $-0.03$ / $+9.56$ \\
      & 1:1 & 31.58 / 54.66 & 31.02 / 60.50 & $-0.56$ / $+5.84$ \\
    \midrule
    \multirow{3}{*}{Muon} & 7:1 & 31.40 / 27.67 & 32.24 / 39.80 & $+0.84$ / $+12.13$ \\
      & 3:1 & 32.16 / 43.67 & 32.05 / 55.19 & $-0.11$ / $+11.52$ \\
      & 1:1 & 31.67 / 59.89 & 31.15 / 65.13 & $-0.52$ / $+5.24$ \\
    \bottomrule
  \end{tabular}}
  \end{center}
\end{table}

Across the six additional pairs, \Method{} improves GSM8K at every ratio by 5.24--12.13 points. Combining these replications with the three \agd{} pairs in Table~\ref{tab:ratio-ablation}, all nine optimizer--ratio settings improve GSM8K. The MMLU shifts follow the same ratio-conditioned pattern under all three optimizers: they are positive at 7:1 ($+0.10$ to $+0.84$), nearly unchanged at 3:1 ($-0.03$ to $-0.11$), and more negative at the reasoning-heavy 1:1 boundary ($-0.52$ to $-0.94$). This shared pattern indicates that the boundary behavior tracks the sampled-data ratio rather than a particular optimizer.

Table~\ref{tab:fixed-weight-optimizer-sweep} places the online rule against the complete fixed-weight grids at K:M $=7{:}1$ for all three optimizers and at $3{:}1$ for Muon. At $7{:}1$, \Method{} improves both benchmarks over direct mixing under every optimizer. In the Muon $3{:}1$ setting, it raises GSM8K by 11.52 points with a 0.11-point MMLU decrease. Its endpoint is bracketed by the fixed $1{:}2$ and $1{:}3$ endpoints on both metrics under Muon at either mixture ratio and under \agd{} at $7{:}1$. Under AdamW, fixed $1{:}3$ is better by 0.24 MMLU points and 2.20 GSM8K points. Thus the online rule does not replace a coefficient selected retrospectively from an oracle grid; it reaches a competitive region without the four additional training runs required by each sweep.

\begin{table*}[ht!]
  \caption{\small Fixed post-warmup loss coefficients and \Method{} across optimizers and K:M mixtures at 30B tokens. Each optimizer uses its learning rate in Table~\ref{tab:hyperparameters}; all other conditions are matched within an optimizer--mixture setting. Scores are final-checkpoint accuracies.}
  \label{tab:fixed-weight-optimizer-sweep}
  \begin{center}
  \resizebox{\textwidth}{!}{%
  \begin{tabular}{lllrrrrrr}
    \toprule
    Optimizer & K:M & Metric & Direct 1:1 & Fixed 1:2 & Fixed 1:3 & Fixed 1:4 & Fixed 1:7 & \Method{} \\
    \midrule
    \multirow{2}{*}{AdamW} & \multirow{2}{*}{7:1} & MMLU & 31.30 & 32.03 & 31.64 & 31.83 & 31.14 & 31.40 \\
      & & GSM8K & 21.91 & 31.24 & 35.86 & 39.20 & 41.39 & 33.66 \\
    \midrule
    \multirow{4}{*}{Muon} & \multirow{2}{*}{7:1} & MMLU & 31.40 & 32.27 & 32.08 & 32.07 & 31.19 & 32.24 \\
      & & GSM8K & 27.67 & 38.59 & 42.46 & 43.29 & 48.98 & 39.80 \\
    \cmidrule(lr){2-9}
      & \multirow{2}{*}{3:1} & MMLU & 32.16 & 32.20 & 31.75 & 31.44 & 30.89 & 32.05 \\
      & & GSM8K & 43.67 & 51.40 & 56.71 & 56.41 & 58.23 & 55.19 \\
    \midrule
    \multirow{2}{*}{\agd{}} & \multirow{2}{*}{7:1} & MMLU & 31.82 & 32.65 & 31.85 & 32.20 & 31.80 & 31.99 \\
      & & GSM8K & 28.58 & 34.19 & 36.39 & 40.18 & 47.31 & 34.50 \\
    \bottomrule
  \end{tabular}}
  \end{center}
\end{table*}

\subsection{Evaluation Settings}
\label{app:evaluation-metrics}

We evaluate general knowledge with MMLU~\citep{hendrycks2021mmlu}, mathematical reasoning with GSM8K~\citep{cobbe2021gsm8k} and MATH~\citep{hendrycks2021math}, and code generation with HumanEval~\citep{chen2021humaneval} and MBPP~\citep{austin2021mbpp}. All evaluations use the complete benchmark split, bfloat16 inference, a maximum context length of 4096, and the settings in Table~\ref{tab:evaluation-metrics}.

\begin{table}[ht!]
  \caption{\small Evaluation protocols. Greedy decoding uses one deterministic completion per problem; Accuracy is the percentage of problems answered correctly or passing all tests.}
  \label{tab:evaluation-metrics}
  \begin{center}
  \resizebox{\linewidth}{!}{%
  \begin{tabular}{lrllll}
    \toprule
    Benchmark & Examples & Prompting & Prediction rule & Max. new tokens & Metric \\
    \midrule
    MMLU & 14,042 & 0-shot & normalized answer-text likelihood & -- & Accuracy \\
    GSM8K & 1,319 & 0-shot & greedy + exact match & 256 & Accuracy \\
    MATH-500 & 500 & 0-shot & greedy + exact match & 256 & Accuracy \\
    HumanEval & 164 & 0-shot instruct & greedy + tests & 400 & Accuracy \\
    MBPP & 500 & 3-shot & greedy + tests & 400 & Accuracy \\
    \bottomrule
  \end{tabular}}
  \end{center}
\end{table}

\paragraph{MMLU.} Every MMLU value in the paper uses the 0-shot FineWeb LightEval recipe~\citep{penedo2024fineweb} over all 57 test subjects. The evaluator scores each complete answer text as a candidate continuation, selects by length-normalized log-likelihood, and reports the micro-averaged \texttt{acc\_norm} over 14,042 questions. This is the MMLU protocol used for both our checkpoints and the Qwen1.5 rows.

\paragraph{GSM8K and MATH.} We use the GSM8K main test split and the MATH-500 test split. Each problem receives one 0-shot greedy completion of at most 256 new tokens, followed by exact matching after benchmark-specific answer extraction and normalization.

\paragraph{HumanEval and MBPP.} HumanEval uses the 0-shot \texttt{humaneval\_instruct} prompt template from lm-evaluation-harness, rendered as plain text without a chat template. MBPP uses its canonical 500-problem test set with three fixed demonstrations. For both benchmarks, the evaluator produces exactly one greedy completion of at most 400 new tokens and executes the benchmark tests; Accuracy is the fraction of problems whose completion passes every test.

Greedy decoding always selects the highest-logit token, with no sampling or multiple candidates. The Qwen1.5 base models use the same task implementations, prompt formats, shot counts, generation limits, normalization rules, and scoring as our final checkpoints, with their native tokenizers and no chat template. All scores are computed once at the final checkpoint and multiplied by 100 for presentation.

\subsection{Training Trajectories at Longer Budgets}
\label{app:long-budget-trajectories}

\begin{figure*}[ht!]
  \centering
  \includegraphics[width=\textwidth]{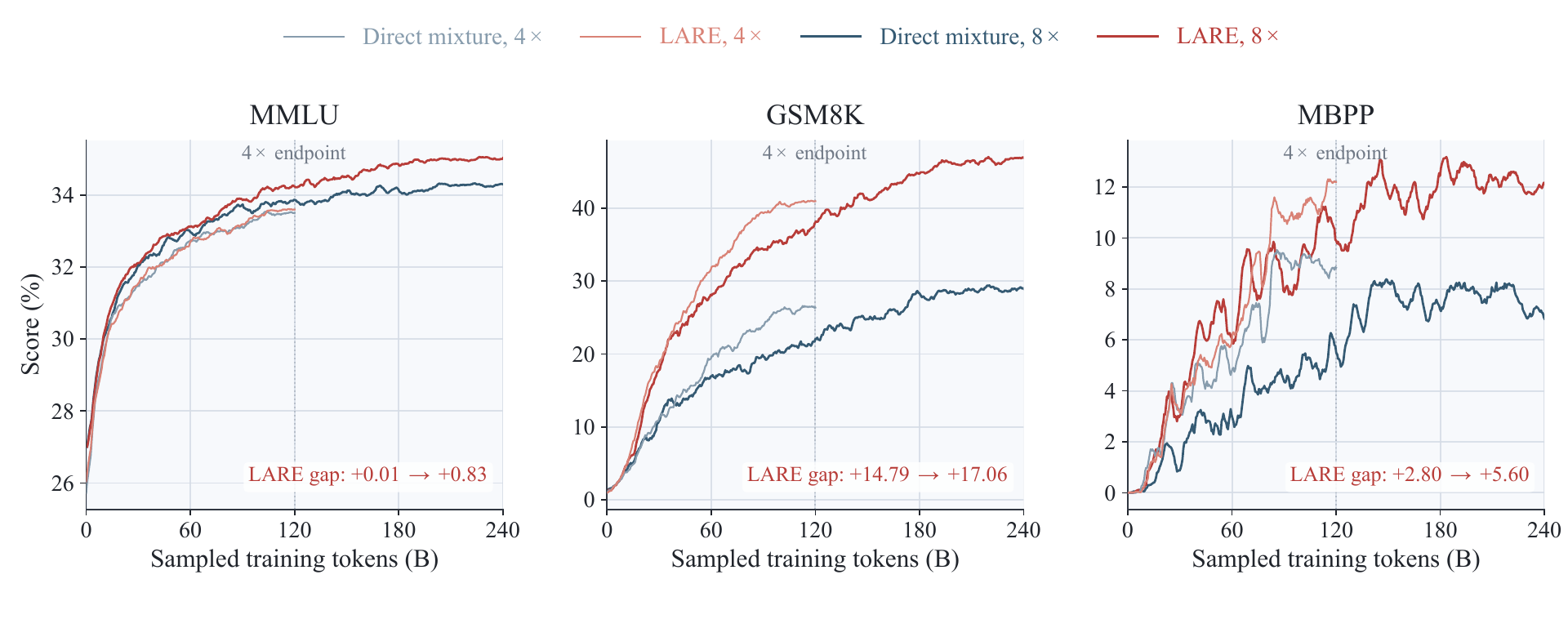}
  \caption{\small\textbf{Checkpoint trajectories for matched $4\times$ and $8\times$ K:M:C runs.} All runs use the K:M:C $=7{:}0.5{:}0.5$ mixture and \agd{}. Blue-gray curves denote direct mixing and red curves \Method{}; lighter curves end at 120B tokens and darker curves at 240B. Curves are centered 11-checkpoint moving averages.}
  \label{fig:main-training-curves}
\end{figure*}

Figure~\ref{fig:main-training-curves} shows checkpoint trajectories for the matched K:M:C $=7{:}0.5{:}0.5$ runs at 120B and 240B sampled tokens.

%% file: sections/94_appendix_scaling/section.tex
\section{Scaling across Model Sizes}
\label{app:scaling}

\subsection{Transferred Setup}
\label{app:scaling-setup}

We test whether the effect of \Method{} transfers across 124M, 350M, and 774M model configurations, using 12/6/768, 24/16/1024, and 36/20/1280 layers/heads/hidden dimensions, respectively. Each model is trained at its nominal $1\times$ Chinchilla allocation~\citep{hoffmann2022training}: 2.52B, 7.08B, and 15.73B sampled tokens. Except for model size, token budget, and the proportionally scaled warmup and \Method{} activation boundary, all training settings follow those used for the main results in Section~\ref{sec:main-results}. All runs use K:M $=7{:}1$. Within each model size, the direct-mixture and \Method{} runs differ only in loss reweighting, while the two stage schedules preserve the same total domain exposure.

\subsection{Replication Results}
\label{app:scaling-results}

\begin{table}[ht!]
\caption{\small\textbf{The mathematical-reasoning benefit of \Method{} emerges and strengthens with joint model--token scale.} Scores are percentages at the final checkpoint under matched $1\times$ Chinchilla budgets. At 774M, \Method{} improves GSM8K by 7.36 points and MMLU by 0.02 points over direct mixing. The 124M result remains at the low-accuracy GSM8K floor, and the multi-stage endpoints remain sensitive to domain order. Bold marks the best result within each model size.}
\label{tab:small-models}
\begin{center}
\begin{tabular}{llccc}
\toprule
Model & Recipe & Sampled tokens & MMLU & GSM8K \\
\midrule
\multirow{4}{*}{124M} & Direct mixture & \multirow{4}{*}{2.52B} & 27.64 & \textbf{2.73} \\
 & K$\rightarrow$M & & 27.82 & 1.59 \\
 & M$\rightarrow$K & & \textbf{27.87} & 0.30 \\
 & \Method{} & & 27.70 & 1.67 \\
\midrule
\multirow{4}{*}{350M} & Direct mixture & \multirow{4}{*}{7.08B} & 28.69 & 3.18 \\
 & K$\rightarrow$M & & 28.14 & 2.27 \\
 & M$\rightarrow$K & & \textbf{28.73} & 1.14 \\
 & \Method{} & & 27.87 & \textbf{5.08} \\
\midrule
\multirow{4}{*}{774M} & Direct mixture & \multirow{4}{*}{15.73B} & 30.28 & 11.37 \\
 & K$\rightarrow$M & & 29.65 & 3.94 \\
 & M$\rightarrow$K & & 29.87 & 0.30 \\
 & \Method{} & & \textbf{30.30} & \textbf{18.73} \\
\bottomrule
\end{tabular}
\end{center}
\end{table}

\begin{figure}[ht!]
\centering
\includegraphics[width=\linewidth]{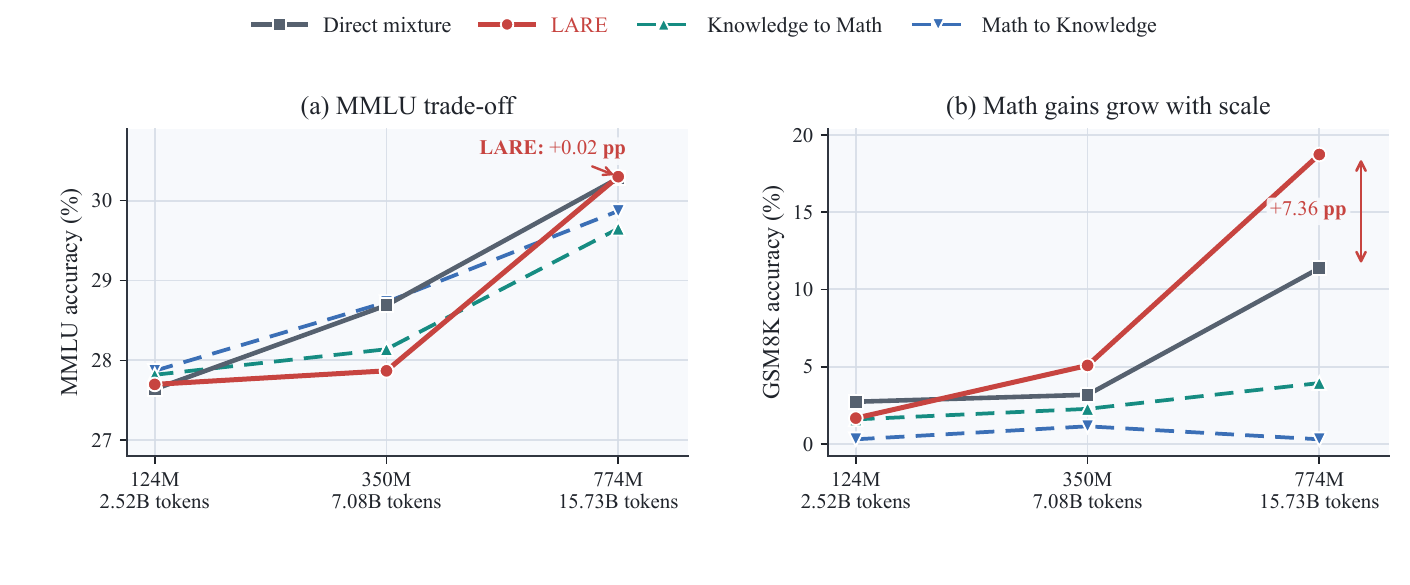}
\caption{\small\textbf{At the largest scale, \Method{}'s growing mathematical-reasoning advantage no longer incurs an MMLU cost.} Curves connect final checkpoints across the matched 124M/2.52B, 350M/7.08B, and 774M/15.73B model--token allocations. Relative to direct mixing, the GSM8K difference changes from $-1.06$ points at 124M to $+1.90$ at 350M and $+7.36$ at 774M; the corresponding MMLU differences are $+0.06$, $-0.82$, and $+0.02$ points. The two ordered stage schedules do not recover the same joint endpoint. Results are single seed and therefore show the observed trend without estimating training-seed uncertainty.}
\label{fig:scaling-replication}
\end{figure}

The validation losses provide a more continuous view of the same effect. Relative to direct mixing, \Method{} reduces mathematics validation loss by 5.70\%, 6.73\%, and 6.55\% at 124M, 350M, and 774M, while increasing knowledge validation loss by only 0.39\%, 0.26\%, and 0.45\%. The downstream GSM8K benefit becomes visible only as the model leaves the low-accuracy regime. Consistently, the final coefficient ratio $w_{\mathrm{M}}/w_{\mathrm{K}}$ selected by \Method{} rises from 2.00 to 2.20 and 2.50 across the three allocations. Because model size and token budget co-vary in this experiment, these results establish a trend under joint Chinchilla scaling rather than isolating parameter count alone.

%% file: sections/96_appendix_agd/section.tex
\section{Activation Gradient Descent}
\label{app:agd}

\agd{} is the base optimizer used in the main experiments, whereas \Method{} is the domain-level objective controller studied in this paper. The two act at different points in the training pipeline: \Method{} determines how strongly each domain contributes before the batch gradient is formed, while \agd{} maps that gradient to a parameter update. We include the derivation here both to make this separation explicit and to specify the optimizer underlying the results in Section~\ref{sec:experiments}.

\subsection{From Activation Matching to a Whitened Update}
\label{app:agd-derivation}

\agd{} is motivated by the view that an optimizer should control the change induced in a layer's activations, rather than only the Euclidean displacement of its parameters. This view parallels layer-wise activation matching in model merging~\citep{xu2025scalable}, but the construction below is local to one optimization step and does not assume an equivalence between model merging and training. More broadly, the observation that different optimization geometries can yield different downstream behavior at similar pretraining loss motivates making this geometry explicit~\citep{chen2026nexuspretraininglossbetter}.

Consider a linear layer applied to $n$ activation vectors. We collect the fixed inputs as $X=[x_1,\ldots,x_n]\in\mathbb{R}^{d_{\mathrm{in}}\times n}$ and the corresponding outputs as $Y=[y_1,\ldots,y_n]=WX$, where $W\in\mathbb{R}^{d_{\mathrm{out}}\times d_{\mathrm{in}}}$. Suppose an ideal local optimizer would change the outputs by $\Delta Y$, while changing the weight matrix by $\Delta W$. Holding $X$ fixed under the local linearization gives
\begin{equation}
(W+\Delta W)X=WX+\Delta WX=Y+\Delta WX.
\label{eq:agd-local-output}
\end{equation}
Consequently, exact activation matching requires $\Delta WX=\Delta Y$. More generally, when one matrix update cannot realize every desired token-level displacement exactly, we choose it by minimizing the residual
\begin{equation}
\min_{\Delta W}\frac{1}{2n}\left\lVert(W+\Delta W)X-(Y+\Delta Y)\right\rVert_F^2
=\min_{\Delta W}\frac{1}{2n}\left\lVert\Delta WX-\Delta Y\right\rVert_F^2.
\label{eq:agd-matching}
\end{equation}

Let $d_i=\nabla_{y_i}\ell_i$ be the gradient of the per-position loss with respect to $y_i$, and collect these vectors as $\Gamma=[d_1,\ldots,d_n]$. The desired output displacement follows the negative-gradient direction, $\Delta Y=-\alpha\Gamma$, where $\alpha>0$ sets its magnitude. Substituting this target into Equation~\ref{eq:agd-matching} gives $\frac{1}{2n}\lVert\Delta WX+\alpha\Gamma\rVert_F^2$. Define the batch gradient and uncentered input second moment as
\begin{equation}
G=\frac{1}{n}\Gamma X^\top
=\frac{1}{n}\sum_{i=1}^{n}d_ix_i^\top,
\qquad
C=\frac{1}{n}XX^\top
=\frac{1}{n}\sum_{i=1}^{n}x_ix_i^\top.
\label{eq:agd-statistics}
\end{equation}
Expanding the squared Frobenius norm then yields the identity
\begin{align}
\frac{1}{2n}\left\lVert\Delta WX+\alpha\Gamma\right\rVert_F^2
={}&\frac{1}{2}\operatorname{Tr}(\Delta WC\Delta W^\top)
+\alpha\langle G,\Delta W\rangle
+\frac{\alpha^2}{2n}\lVert\Gamma\rVert_F^2.
\label{eq:agd-surrogate}
\end{align}
The last term is independent of $\Delta W$, and the scalar $\alpha$ can be absorbed into the eventual learning rate. Setting it to one therefore recovers the activation-matching surrogate $\Phi(\Delta W)=\langle G,\Delta W\rangle+\frac{1}{2}\operatorname{Tr}(\Delta WC\Delta W^\top)$ used below.

The role of $C$ is important when the activation spectrum is long-tailed. In raw parameter coordinates, a component of $\Delta W$ aligned with a high-variance eigenvector of $C$ produces a much larger output displacement than an equally sized component aligned with a low-variance direction. A Euclidean constraint on $\Delta W$ therefore allows a few high-variance activation directions to dominate the matching objective. To remove this scale imbalance, we whiten the inputs and express the update in the corresponding parameter coordinates. Assuming $C$ is positive definite for the derivation, with numerical regularization used in practice, define
\begin{equation}
\widetilde X=C^{-1/2}X,
\qquad
\frac{1}{n}\widetilde X\widetilde X^\top=I,
\qquad
Z=\Delta WC^{1/2},
\qquad
M=GC^{-1/2}.
\label{eq:agd-whitening}
\end{equation}
Here $Z$ is precisely the change to the layer weight expressed in the whitened input basis, since $WX=(WC^{1/2})\widetilde X$. The change of variables gives
\begin{equation}
\langle G,\Delta W\rangle=\langle M,Z\rangle,
\qquad
\operatorname{Tr}(\Delta WC\Delta W^\top)=\lVert Z\rVert_F^2.
\label{eq:agd-whitening-identities}
\end{equation}
Thus whitening converts the activation metric into an isotropic Frobenius metric and prevents the long tail of the covariance spectrum from directly setting the update scale. The unconstrained minimizer is $Z=-M$, which maps back to $\Delta W=-GC^{-1}$. To control the structure of the direction independently of its global magnitude, which is supplied by the learning rate, we instead solve the linear minimization oracle
\begin{equation}
Z^\star\in\argmin_{\lVert Z\rVert_{\mathcal{B}}\leq 1}\langle M,Z\rangle,
\qquad
\Delta W^\star=Z^\star C^{-1/2}.
\label{eq:agd-lmo}
\end{equation}
The constraint is therefore imposed in the whitened activation geometry rather than directly on the raw parameter displacement.

\subsection{The Row-Wise \texorpdfstring{$(2,\infty)$}{(2,infty)} Direction}
\label{app:agd-rowwise}

We use $\lVert Z\rVert_{2,\infty}=\max_j\lVert Z_j\rVert_2$, where $Z_j$ denotes output row $j$. The corresponding oracle decomposes over rows and yields
\begin{equation}
(\Delta W_j)^\star
=-
\frac{G_jC^{-1}}
{\sqrt{G_jC^{-1}G_j^\top}+\epsilon},
\qquad j=1,\ldots,d_{\mathrm{out}}.
\label{eq:agd-update}
\end{equation}
Ignoring the numerical constant, each row satisfies $(\Delta W_j)^\star C((\Delta W_j)^\star)^\top=1$. Hence every output row receives a comparable step in the activation metric even when the raw gradient norms or input variances differ substantially. The normalization is local to each output row, so a small number of large rows cannot set the scale of the entire matrix update.

Table~\ref{tab:agd-constraint-directions} summarizes the closed-form directions induced by the three constraint choices used in our optimizer family. Here $\lVert\cdot\rVert_\infty$ is the elementwise maximum norm, $\lVert\cdot\rVert_2$ is the spectral norm, and $\operatorname{Orth}(\cdot)$ denotes the polar factor.

\begin{table*}[ht!]
\caption{Closed-form ActGD directions under three constraints in the whitened activation geometry, where $M=GC^{-1/2}$ and $\Delta W^\star=Z^\star C^{-1/2}$. The main experiments use $(2,\infty)$; implementations add $\epsilon$ stabilization where needed.}
\label{tab:agd-constraint-directions}
\begin{center}
\resizebox{\textwidth}{!}{%
\begin{tabular}{lll}
\toprule
Constraint on $Z$ & Whitened direction $Z^\star$ & Parameter-space update $\Delta W^\star$ \\
\midrule
$\lVert Z\rVert_{2,\infty}\leq 1$ & $Z_j^\star=-M_j/\lVert M_j\rVert_2$ & $(\Delta W_j)^\star=-G_jC^{-1}/\sqrt{G_jC^{-1}G_j^\top}$ \\
$\lVert Z\rVert_\infty\leq 1$ & $Z^\star=-\operatorname{sign}(M)$ & $\Delta W^\star=-\operatorname{sign}(GC^{-1/2})C^{-1/2}$ \\
$\lVert Z\rVert_2\leq 1$ & $Z^\star=-\operatorname{Orth}(M)$ & $\Delta W^\star=-\operatorname{Orth}(GC^{-1/2})C^{-1/2}$ \\
\bottomrule
\end{tabular}}
\end{center}
\end{table*}

We adopt Equation~\ref{eq:agd-update}: unlike the elementwise sign direction, it preserves within-row gradient structure, and unlike the spectral direction, it normalizes each output row independently rather than coupling all rows through a single matrix-wide polar factor.

\subsection{Stateful Optimizer Used in Our Experiments}
\label{app:agd-state}

The practical optimizer smooths both quantities in Equation~\ref{eq:agd-statistics}. Let $G_t$ and $C_t$ denote their step-$t$ estimates. For gradient-momentum decay $\beta_g$ and covariance decay $\beta_c$, we maintain
\begin{align}
H_t&=\beta_gH_{t-1}+(1-\beta_g)G_t,
&\widehat H_t&=\frac{H_t}{1-\beta_g^t},\nonumber\\
S_t&=\beta_cS_{t-1}+(1-\beta_c)C_t,
&\widehat S_t&=\frac{S_t}{1-\beta_c^t}.
\label{eq:agd-ema}
\end{align}
Let $A_t$ be a numerically stabilized Newton--Schulz approximation to $\widehat S_t^{-1}$ and set $U_t=\widehat H_tA_t$. The implemented descent transform is
\begin{equation}
D_{t,j}=\frac{U_{t,j}}{\sqrt{\max\{0,\widehat H_{t,j}U_{t,j}^\top\}}+\epsilon},
\qquad
W_{t+1}=(1-\eta_t\lambda)W_t-\eta_tD_t,
\label{eq:agd-practical-update}
\end{equation}
where $\lambda$ is decoupled weight decay. The clamp affects only roundoff-induced negative quadratic forms. Equations~\ref{eq:agd-ema}--\ref{eq:agd-practical-update} reduce to the row-wise activation-space direction in Equation~\ref{eq:agd-update} when the running estimates and inverse approximation are exact.


Table~\ref{tab:actgd-settings} records the \agd{}-specific settings shared by the main experiments. The token embedding, language-model head, and transformer-block matrices use the same row-wise update family and the same hyperparameter values, although they are retained as two parameter groups for bookkeeping. This unified treatment is the single-configuration property highlighted in Table~\ref{tab:optimizer-ablation}.

\begin{table}[t]
\caption{\small\textbf{ActGD uses one optimizer configuration for both vocabulary-facing and transformer-block parameters in the main experiments.} The settings below determine the update geometry; the common model, batch, and learning-rate schedule are reported in Appendix~\ref{app:hyperparameters}.}
\label{tab:actgd-settings}
\begin{center}
\begin{tabular}{ll}
\toprule
Setting & Value \\
\midrule
Embedding/head update & Row-wise $(2,\infty)$ \\
Transformer-block update & Row-wise $(2,\infty)$ \\
Peak learning rate $\eta$ & 0.01 for both groups \\
Gradient/covariance EMA $(\beta_g,\beta_c)$ & $(0.95,0.95)$ \\
Numerical constant $\epsilon$ & $10^{-5}$ \\
Newton--Schulz iterations & 4 \\
Covariance sample cap & $n'=\min(n,d_{\mathrm{in}})$ \\
Decoupled weight decay $\lambda$ & $10^{-5}$ \\
\bottomrule
\end{tabular}
\end{center}
\end{table}

\subsection{Parameter Coverage and Distributed Implementation}
\label{app:agd-implementation}

For a linear map, forward hooks flatten the leading activation dimensions and estimate $C_t=X_t^\top X_t/n$ in FP32. When the number of activation rows exceeds the input width, the main configuration samples $n'=d_{\mathrm{in}}$ rows with replacement before forming the covariance. Query, key, and value projections in the same attention block consume the same input tensor, so their covariance is computed and communicated once and then reused for all three weights. The covariance EMA and the gradient EMA are stored in FP32; the inverse is evaluated with four fixed Newton--Schulz polynomial iterations.

The embedding layer is the one-hot special case of the same construction. Its input second moment is diagonal, so we store only the empirical token-frequency vector and apply its inverse by row-wise scaling in the stored embedding orientation rather than materializing a vocabulary-sized covariance matrix. The language-model head is treated as an ordinary linear layer using the second moment of its hidden-state inputs. Because the architecture used here has bias-free linear maps and parameter-free normalization, all learned parameters are matrices and no separate one-dimensional fallback optimizer is required.

In distributed training, activation statistics are averaged across data-parallel ranks before the update is constructed. Each parameter's inverse and update are computed on one deterministically assigned owner rank, after which the disjoint update shards are combined with a sum reduction. The communicated update buffer uses bfloat16, while the persistent EMA states remain FP32. The sharded optimizer checkpoints store the step counter and both EMA tensors for each owned parameter; forward-hook statistics, shared-covariance aliases, and temporary inverse buffers are step-local and are cleared after each update.

ActGD stores a gradient-sized first-moment tensor and an input-covariance tensor for each matrix, and its additional computation is dominated by covariance formation and the approximate inverse. Activation subsampling, query/key/value covariance reuse, cached identity matrices, and distributed owner sharding reduce this overhead without changing Equation~\ref{eq:agd-practical-update}. Since the experiments in this paper match sampled tokens rather than wall-clock time, the comparisons in Tables~\ref{tab:optimizer-ablation} and~\ref{tab:main-results} should be interpreted as optimization-quality and sample-efficiency comparisons, not as claims of equal throughput or equal peak memory.